\documentclass[journal]{IEEEtran}
\IEEEoverridecommandlockouts
\usepackage[noadjust]{cite} 
\usepackage{amsmath, amssymb, amsfonts}
\usepackage{graphicx}
\usepackage{textcomp}
\usepackage{xcolor}
\def\BibTeX{{\rm B\kern-.05em{\sc i\kern-.025em b}\kern-.08em
    T\kern-.1667em\lower.7ex\hbox{E}\kern-.125emX}}

\usepackage{mathtools}
\usepackage{hyperref}
\usepackage{multirow}

\usepackage[labelformat=simple]{subcaption}

\input{mysymbol.sty}
\usepackage{siunitx}
\usepackage[normalem]{ulem}
\usepackage{booktabs}

\usepackage{cleveref} 

\usepackage[colorinlistoftodos]{todonotes}

\newlength\myindent 
\def \qdata { q_{\mathrm{data}} }

\def \pgen { p_{\bbtheta} }
\def \pgenstar { p_{\bbtheta^\star} }

\def \alphacumprod { \bar{\alpha} }

\def \enc { \bbPhi^{\rmE} }
\def \dec { \bbPhi^{\rmD} }
\def \bneck { \bbPhi^{\rmB} }

\def \thetaenc { \Theta^{\rmE} }
\def \thetadec { \Theta^{\rmD} }
\def \thetabneck { \Theta^{\rmB} }

\def \sigenc { \bbZ }
\def \sigdec { \bbY }

\def \projlayer { \bbPi }
\def \projlayercond { \projlayer^{\rmU} }
\def \projlayerdiffusionstep { \projlayer^{\rmk} }
\def \projlayerskip { \projlayer^{\mathrm{skip}} }

\def \Th { T_{h} }
\def \Tp { T_{p} }

\newcommand{\uday}[1] {\bbu^{(#1)}}
\newcommand{\rday}[1] {\bbr^{(#1)}}

\newcommand{\subjectto}{\text{subject~to}}

\def \svpos { \vspace{.2em} }

\def \algenv {1}

\ifnum \algenv=0
    \usepackage{algorithmic}
    \usepackage{algorithm}
\else 
    \ifnum \algenv=1

        \usepackage{algorithm}
        \usepackage{algpseudocode}
        \usepackage{float} 
        
        \algrenewcommand\algorithmiccomment[1]{\hfill$\triangleright$~#1}
        
        \algnewcommand{\LineComment}[1]{\Statex $/*$~#1~$*/$}

        \algnewcommand{\LineCommentIndented}[1]{\Statex \hspace{\algorithmicindent} $/*$~#1~$*/$}

    \else 
        \usepackage{algorithm2e}
    \fi
\fi

\usepackage{setspace} 

\newif\ifcompiletikz
\compiletikztrue
\compiletikzfalse

\usetikzlibrary{arrows.meta,positioning} 
\usetikzlibrary{backgrounds,calc}

\usepackage{amsthm}

\newtheorem{remark}{\hspace{0pt}\bf Remark}

\usepackage{listings}
\usepackage{enumerate}
\usepackage{enumitem}
\usepackage{bbm}

\def\x{{\mathbf x}}

\date{\today}

\def\E{\mathbb{E}}

\usepackage{caption}
\graphicspath{
    {./figures/}
    {./figures/sp500/sociable-frigatebird-619/}
}

\usepackage{tikz}
\usetikzlibrary{external}
\usetikzlibrary{arrows.meta, positioning, calc, backgrounds, fit}

\definecolor{gnncol}{RGB}{200,230,190}
\definecolor{projcol}{RGB}{223,201,234}
\definecolor{selcol}{RGB}{250,205,158}
\definecolor{bandcol}{RGB}{176,130,80}   

\providecommand{\bbS}{\mathbf{S}}
\providecommand{\bbC}{\mathbf{C}}
\providecommand{\bbD}{\mathbf{D}}
\providecommand{\bbZ}{\mathbf{Z}}
\providecommand{\bbY}{\mathbf{Y}}
\providecommand{\bbU}{\mathbf{U}}
\providecommand{\bbK}{\mathbf{K}}
\providecommand{\bbP}{\mathbf{P}}
\providecommand{\bbV}{\mathbf{V}}
\providecommand{\bbI}{\mathbf{I}}

\providecommand{\bbx}{\mathbf{x}}
\providecommand{\bbu}{\mathbf{u}}
\providecommand{\bbeps}{\boldsymbol{\epsilon}_{\boldsymbol{\theta}}}

\providecommand{\thetaenc}{\boldsymbol{\Theta}^{\mathrm{enc}}}
\providecommand{\thetadec}{\boldsymbol{\Theta}^{\mathrm{dec}}}
\providecommand{\thetasel}{\boldsymbol{\Theta}^{\mathrm{sel}}}

\ifdefined\ifshowcaption\else \newif\ifshowcaption \showcaptiontrue \fi
\ifdefined\nocaption\showcaptionfalse\fi

\showcaptionfalse   

\usepackage[most]{tcolorbox}   

\begin{document}

\title{Generative Diffusion Models of Stochastic \\Graph Signals}

\author{Yi\u{g}it~Berkay~Uslu, Samar~Hadou, Sergio~Rozada, Shirin~Saeedi~Bidokhti~\IEEEmembership{Member,~IEEE,} ~Alejandro~Ribeiro,~\IEEEmembership{Fellow,~IEEE}

\thanks{
Preliminary results of this work were presented in part at the ICASSP 2026 conference~\cite{uslu2026graph} and appear in the preprint~\cite{uslu2026graphsignalwireless}.
Y. Berkay Uslu, Samar Hadou, Shirin S. Bidokhti and Alejandro Ribeiro are with the Dept. of Electrical and Systems Eng., University of Pennsylvania, Philadelphia, PA 19104 USA.
}
\thanks{
Sergio Rozada is with the Department of Signal Processing and Communications, King Juan Carlos University, Madrid, Spain.
}

\thanks{The implementation code is available at \href{https://github.com/yigit-uslu/graph-signal-diffusion-modeling}{\url{https://github.com/yigit-uslu/graph-signal-diffusion}}.}

}

\maketitle
\thispagestyle{plain}
\pagestyle{plain}

\begin{abstract}
Sampling stochastic signals supported on a graph underlies many graph machine learning tasks, including recommender systems, forecasting in financial markets, and wireless network optimization. In these settings, the target signals are realizations of unknown conditional distributions. However, prevailing approaches rely mostly on intricate, application-tailored designs that often regress to a conditional mean instead of sampling from the conditional law. This paper unifies such problems as conditional graph signal generative modeling and tackles them with a single denoising diffusion framework. We learn a reverse diffusion process, parametrized by graph neural networks (GNNs), that draws graph signals conditioned directly on the graph topology and on node-feature side information. The reverse process is realized by a novel architecture, the U-Graph Neural Network (U-GNN), which generalizes the image-convolutional U-Net to graph-structured signals. The U-GNN performs multi-resolution encoder--decoder processing in which pooling and unpooling reduce to a learned node selection, expressed by nested selection matrices, and a zero-padded lifting of coarse signals back to the full node set. The graph convolutions are carried out on the original graph, with a stride that sets their hop reach, so the U-GNN bypasses explicit graph coarsening at every resolution. 
We demonstrate our method on two generative tasks: stock price forecasting and optimal wireless resource allocation, with extensive numerical results in both domains.
\end{abstract}

\section{Introduction}

Signals defined over irregular, graph-structured domains are pervasive, spanning application areas such as recommender systems \cite{wang2019neural, mao2021ultragcn, liu2023personalized}, wireless communications \cite{eisen2020optimal, shen2023graph, zhao2023linkscheduling, wang2024engnn, chowdhury2024deepunfolding, uslu2025faststateaugmentedlearningwireless}, %
and forecasting in financial markets \cite{chen2018incorporating, sawhney2020spatiotemporal, uddin2023attention}. %
In many such settings, these signals are realizations from unknown probability distributions rather than deterministic quantities \cite{perraudin2017stationary}. 
This paper proposes a generative modeling framework that learns to sample from such graph signal distributions.Generative modeling has proved effective for synthesizing graph topologies themselves~\cite{jo2022score, vignac2023digress}, with molecular generation serving as a canonical example~\cite{hoogeboom2022equivariant}. We instead take the graph structure as given and learn to generate node-defined signals.

A wide array of generative paradigms exists for learning to sample from unknown distributions \cite{kingma2014auto, goodfellow2014generative, rezende2015variational}, such as variational autoencoders, generative adversarial networks, and normalizing flows. Among them, denoising diffusion models and their score- and flow-based variants stand out for their sample quality, training stability, and principled conditional inference in high dimensions \cite{ho2020denoising, song2021denoising, song2021score, lipman2023flow}. While these models differ in what the parametric model approximates and in the sampling dynamics, they share a common structure. A fixed forward process gradually corrupts the data into a tractable reference prior, and a learned reverse process denoises samples from that prior back toward the data distribution. Graph signals have recently become a target for generative, particularly diffusion, modeling in domains such as the forecasting of traffic and financial series~\cite{uslu2026graph, wen2023diffstg, daiya2024diffstock, chen2025dhmoe, rozada2026graphaware}
, and wireless optimization~\cite{darabi2024diffusion, diffsg2024liang, uslu2025generativediffusionmodelsresource, guo2025diffusionmultipleantenna, zhizhou2025scorebasedrismimo, uslu2026graphsignalwireless, lee2026userspecificchannel}%
. However, these efforts have mostly been application-specific, with task-tailored denoiser architectures. Their formulations also often remain deterministic, seeking solution distributions whose mass concentrates near an optimal deterministic solution rather than drawing diversified samples from the conditional law.


In this paper, we propose a general methodological framework for graph signal generation. 
At its core, this requires designing a diffusion model whose denoising network encodes inductive biases tailored to the data domain. For images, the standard denoising backbone is the U-Net with a symmetric encoder--decoder structure and multi-resolution hierarchy
\cite{ronneberger2015unet, nnunet2021isensee}
. The U-Net is built on convolutional neural networks (CNNs), whose layers apply translation-equivariant filters on the regular pixel grid. The natural
counterpart for graph-structured data is the graph neural network (GNN), whose layers apply graph-shift-equivariant filters, thereby extending convolutional processing to irregular domains~\cite{gama2018convolutional}. Replacing CNNs with GNN layers does not, on its own, suffice to adapt the U-Net design to the denoising of graph signals. The main obstacle is the pooling stage. In the image setting, the pixel grid provides a regular pooling lattice that persists after each down-sampling step, thus, feature extraction and spatial reduction compose seamlessly. Graph data lacks such a canonical lattice. A common workaround coarsens the graph at each resolution through a learned pooling operator, as in differentiable cluster-assignment pooling and top-$k$ node-scoring schemes \cite{ying2018hierarchical, gao2019graph}
. These methods are mainly designed for graph-level tasks, such as classification. Under arbitrary coarsening, however, the pooled signal loses a clear interpretation as a convolutional signal on the original graph.

We take a different route in this paper. We formalize down- and up-sampling, equivalently graph or node (un)pooling, as a \emph{learned node (de)selection} operation, expressed through nested selection matrices \cite{gama2018convolutional}. These matrices are trained end-to-end with the denoising network, so each resolution retains the nodes most informative for the denoising task at hand. The selection operator and its transpose play two complementary roles. First, the nested hierarchy of selections instantiates the multi-resolution encoder--decoder cascade. Each encoder selection picks a smaller active node subset that defines the next coarser resolution, and the matching transpose in the decoder lifts coarse signals back to the finer one. Second, at a fixed resolution, the same operator pair keeps GNN filtering interpretable as a convolution on the original graph. The transpose zero-pads a node-reduced signal to the full vertex set, the GNN layer filters there, and the selection returns the result to the reduced domain. Unlike the coarsening schemes above, filtering therefore stays convolutional on the original graph at every resolution. We further introduce a stride parameter that sets the hop distance of each graph shift. This mirrors how dilated convolutions in CNNs replace fixed pooling rules, folding down-sampling and receptive-field growth into the convolution itself \cite{springenberg2015striving, yu2016multiscale}
. On the active nodes, a larger stride enlarges the aggregation neighborhoods and prevents the graph filters from drawing mostly from zero-padded positions under heavy down-sampling. This construction avoids explicit graph coarsening and lets convolutional filtering and pooling compose across arbitrary resolutions.

Building on this mechanism, we propose the \emph{U-Graph Neural Network (U-GNN)} architecture as a graph-domain adaptation of the classical U-Net. It comprises a U-shaped
encoder--decoder pipeline with skip connections, built from GNN layers and the pooling and unpooling operators
above. We then cast graph signal generation as a conditional denoising diffusion process and parametrize its reverse process with a U-GNN. The generations are thereby conditioned directly on the graph topology and on node-feature side information. Although we focus on diffusion denoising, the U-GNN is a general-purpose architecture for learning over graph signals. It inherits the permutation equivariance, stability to perturbations, and cross-topology transferability of GNNs \cite{ruiz2021graph, shen2023graph}.

We demonstrate the framework and the U-GNN architecture on two generative tasks. The first is stock price forecasting over a correlation graph of financial indicators. Given recent market history, we generate future
price trajectories as graph signals, capturing the market uncertainty and rare tail events that point forecasts and statistical baselines often miss. The second is resource allocation in wireless networks under ergodic
quality-of-service (QoS) requirements, where optimal policies are inherently
probabilistic and are realized through time-sharing. Treating solution samples from an expert primal--dual algorithm as graph signals on channel (interference) graphs, we learn to sample from the expert allocation distribution given channel and user-state information.

A brief summary of our contributions is as follows.
\begin{enumerate}[leftmargin=1cm, itemindent=0.0cm, label = \textbf{(C\arabic*).}]
    \item We develop a graph signal generative modeling approach that combines
    denoising diffusion models with GNNs to sample stochastic graph signals.
    \svpos

    \item We parametrize the reverse diffusion process with a novel U-GNN
    architecture, whose building blocks are GNN layers with stride and
    zero-padded pooling, that generalizes image-convolutional U-Nets to graph-structured signals.
    \svpos

    \item We apply our approach to stock price forecasting and optimal wireless resource allocation, with extensive numerical results in both domains. 

\end{enumerate}

The rest of the paper is organized as follows.
Section~\ref{sec:graph-signal-generative-models} presents the conditional graph signal generative diffusion framework.
Section~\ref{sec:gnns} introduces GNNs with built-in stride and pooling. Section~\ref{sec:ugnn-proposed} assembles the U-GNN architecture. Section~\ref{sec:experiments} validates the method on wireless resource allocation and financial forecasting. Section~\ref{sec:conclusion} concludes the paper.

\section{Graph Signal Generative Diffusion Models}
\label{sec:graph-signal-generative-models}

Consider a weighted graph $\ccalG = (\ccalV, \ccalE, \ccalW)$ on $N$ nodes, with node set $\ccalV = \{1, \ldots, N\}$, edge set $\ccalE \subseteq \ccalV \times \ccalV$, and weight map $\ccalW : \ccalE \to \reals$. A graph signal assigns features to the nodes and is collected in a matrix $\bbX = [\bbx^{1}, \ldots, \bbx^{F}] \in \reals^{N \times F}$, whose $f$th column $\bbx^{f} \in \reals^N$ gathers the $f$th feature across all nodes. We endow $\ccalG$ with a graph shift operator (GSO) $\bbS \in \reals^{N \times N}$, a sparse matrix with $[\bbS]_{m,n} = 0$ whenever $m \neq n$ and $(m, n) \notin \ccalE$. We note that this GSO definition allows the diagonal entries to be chosen freely. The GSO encodes the graph topology, and successive products (shifts) $\bbS \bbX$ propagate signal values along the edges. Common GSO choices are the (weighted) adjacency matrix $\bbA$ and the graph Laplacian $\bbL = \mathrm{diag}(\bbA \bbone_N) - \bbA$.

We address the conditional generation of graph signals. The target is a graph signal $\bbx_0 \in \reals^N$ on a known graph $\ccalG$. The conditioning collects the topology, through the GSO $\bbS$, and auxiliary node states $\bbu \in \reals^{N \times U}$ (side information) into the variable $\bbxi = (\bbS, \bbu)$. Let $q_{\bbxi}$ denote the distribution of the conditioning and $q_{\bbx \vert \bbxi}(\cdot \cond \bbxi)$ the law of the target given $\bbxi$. These define the joint data distribution $\qdata = q_{\bbx \vert \bbxi}\, q_{\bbxi}$, from which we draw pairs $(\bbx_0, \bbxi)$. We seek a parametric generator $\pgenstar(\cdot \cond \bbxi) = p(\cdot \cond \bbxi; \bbtheta^*)$ whose conditional law matches $q_{\bbx \vert \bbxi}(\cdot \cond \bbxi)$ for $q_{\bbxi}$-almost every $\bbxi$, i.e.,
\begin{align} \label{eq:generative-modeling}
    \bbtheta^\star \in \argmin_{\bbtheta} \E_{\bbxi \sim q_{\bbxi}} \big[ \kldiv[\big]{q_{\bbx \vert \bbxi}(\cdot \cond \bbxi)}{\pgen(\cdot \cond \bbxi)} \big].
\end{align}
The conditional laws are unknown in closed form. We assume samples $(\bbx_0, \bbxi) \sim \qdata$ are available. Therefore, we realize the optimal generator $\pgenstar$ as a conditional denoising diffusion model, trained on a dataset $\{ (\bbx^{(i)}_0, \bbxi^{(i)}) \}_{i=1}^{|\ccalD|} \sim \qdata$.

\subsection{Denoising Diffusion Models for Graph Signals}


\compiletikzfalse

\newif \ifdrawbands
\drawbandsfalse 

\begin{figure*}[t]

\ifcompiletikz

\else

    \includegraphics[width=\linewidth]{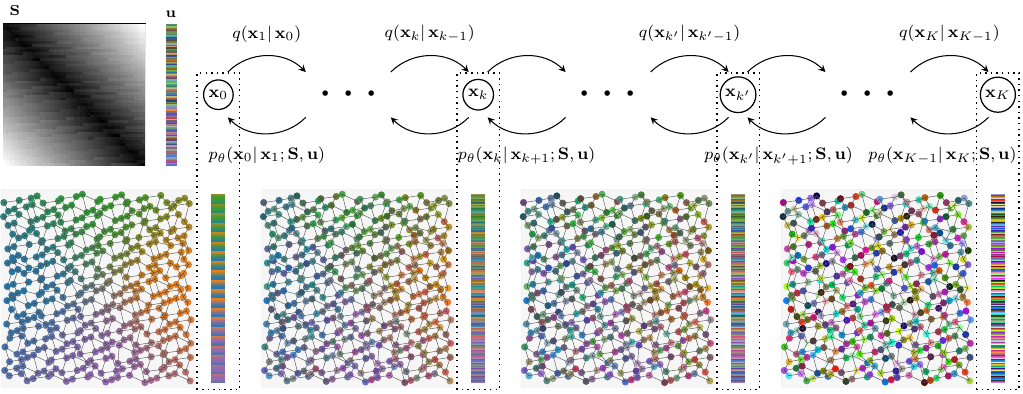}

\caption{\textbf{A denoising diffusion model of graph signals.} 
A forward noising process $q(\bbx_k \cond \bbx_{k-1})$ gradually removes structure from the original graph signals by adding white noise, with $\bbx_K \approx \mathcal{N}(\bb0, \bbI)$. A denoising diffusion process $\pgen(\bbx_{k-1} \vert \bbx_k; \bbS, \bbu)$ is trained to reverse the forward process and generate novel graph signal samples $\bbx_0$ distributed approximately by the conditional distributions $\qdata(\bbx_0 \cond \bbS, \bbu)$ for a given GSO $\bbS$ and observation values represented by another graph signal $\bbu$.
}

\label{fig:diffusion_process_training}

\end{figure*}


A denoising diffusion model pairs a fixed forward process $q(\bbx_{1:K} \cond \bbx_0)$, which noises clean targets toward a tractable reference prior, with a learned reverse process $\pgen(\bbx_{k-1} \cond \bbx_k)$, which denoises prior samples back toward the data. Conditional generation requires no additional machinery. We condition the reverse process on $\bbxi$ as $\pgen(\bbx_{k-1} \cond \bbx_k;\, \bbxi)$, leaving the forward process and the training objective unchanged. Fig.~\ref{fig:diffusion_process_training} illustrates this construction, which we detail next.

\smallskip
\noindent \textbf{Forward noising process.} 
The forward process is a Markov chain that adds noise to the target $\bbx_0$ over $K$ steps. Given $(\bbx_0, \bbxi) \sim q_0 = \qdata$ with $\bbxi$ held fixed, we corrupt $\bbx_0$ with white noise on an increasing schedule $\{ \beta_k \}_{k=1}^{K}$ (e.g., linear), producing iterates $\{ \bbx_k \}_{k=1}^{K}$ through the Gaussian transitions,
\begin{align} \label{eq:forward-diffusion:transition-kernel}
    q( \bbx_k \cond \bbx_{k-1} ) = \ccalN( \bbx_k; \sqrt{1 - \beta_k}\, \bbx_{k-1}, \beta_k \bbI).
\end{align}
Since the drift is linear and the noise increments are independent, the conditionals remain Gaussian at each step $k$, i.e.,
\begin{align} \label{eq:forward-diffusion:conditionals}
    q_k(\bbx_k \cond \bbx_0) = \mathcal{N} \big( \bbx_k;\, \sqrt{\alphacumprod_k}\, \bbx_0,\, (1 - \alphacumprod_k)\, \bbI \big),
\end{align}
with $\alpha_k \coloneqq 1 - \beta_k$ and $\alphacumprod_k \coloneqq \prod_{m=1}^k \alpha_m$ for $k = 1, \ldots, K$. A standard reparametrization of \eqref{eq:forward-diffusion:conditionals} expresses $\bbx_k$ as a deterministic map of $(\bbx_0, k)$ plus independent standard noise~\cite{ho2020denoising},
\begin{align}\label{eq:reparametrization-trick}
    \hspace{-.25em}\bbx_k(\bbx_0, k, \bbepsilon) \!=\! \sqrt{ \alphacumprod_k }\, \bbx_0 \!+\! \sqrt{1 - \alphacumprod_k }\, \bbepsilon, \quad \bbepsilon \sim \mathcal{N}(\bb0, \bbI) \!\perp\! \bbx_0,
\end{align}
and it permits drawing any iterate $\bbx_k \sim q_k(\cdot \cond \bbx_0, \bbxi)$ directly, without simulating the chain in \eqref{eq:forward-diffusion:transition-kernel}. For a suitable schedule and a large enough $K$, the process drives $\bbx_0 \sim q_0(\cdot \cond \bbxi)$ toward a Gaussian prior, with $\bbx_K \sim q_K \approx \mathcal{N}(\bb0, \bbI)$ for every $\bbxi$.

\smallskip
\noindent \textbf{Backward (reverse) denoising process.}
Denoising diffusion fixes $p_K(\cdot \cond \bbxi) = q_K \approx \mathcal{N}(\bb0, \bbI)$ and learns a reverse Markov chain whose kernels $\pgen(\bbx_{k-1} \cond \bbx_k; \bbxi)$ approximate the time reversal of the forward chain. Consistent with the continuous-time limit, we model each reverse step as Gaussian transitions,
\begin{align} \label{eq:reverse-Gaussian-kernel}
   \hspace{-0.25em} p_{\bbtheta}(\bbx_{k-1} \!\given\! \bbx_k; \bbxi) \!=\! \mathcal{N}\!\left( \bbx_{k-1}; \bbmu_{\bbtheta}\big( \bbx_k, k; \bbxi \big), \bbSigma_{\bbtheta} \big( \bbx_k, k; \bbxi \big) \right).
\end{align}
As is standard, we fix the covariance $\bbSigma_{\bbtheta} \!=\! \sigma_k^2 \bbI$ for a predefined $\sigma_k$ tied to the schedule $\beta_k$ and learn only the mean $\bbmu_{\bbtheta}$.

Training maximizes the evidence lower bound (ELBO) on the log-likelihood $\log \pgen(\cdot \given \bbxi)$. Under the fixed covariance and the reparametrization \eqref{eq:reparametrization-trick}, this is equivalent to a noise-prediction objective. Omitting its per-step ELBO weights matches, and often improves upon, 
the weighted objective in practice \cite{ho2020denoising}. We therefore minimize the following objective,
\begin{align} \label{eq:ddpm-noise-training}
    \bbtheta^\star \in \argmin_{\bbtheta} \ccalL(\bbtheta) \coloneqq \E \Big[ \big\| \bbepsilon - \bbepsilon_{\bbtheta} \big( \bbx_k(\bbx_0, k, \bbepsilon), k; \bbxi \big) \big\|^2 \Big],
\end{align}
where the expectation is over $(\bbx_0, \bbxi) \sim \qdata$ and diffusion steps $k \sim \mathrm{unif}\{1, \ldots, K \}$. 

By the reparametrization \eqref{eq:reparametrization-trick}, the trained noise predictor $\bbepsilon_{\bbtheta^\star}(\bbx_k, k; \bbxi)$ yields both a \emph{graph signal denoiser} and the mean of the reverse kernel in \eqref{eq:reverse-Gaussian-kernel},
\begin{align}
    \widehat{\bbx}_{0}(\bbx_k, k; \bbxi) &= \frac{\bbx_k - \sqrt{1 - \alphacumprod_k}\, \bbepsilon_{\bbtheta^\star} }{\sqrt{\alphacumprod_k}}, \label{eq:noisepred-denoiser-param-equivalence} \\
    \bbmu_{\bbtheta^\star}(\bbx_k, k; \bbxi) &= \frac{1}{ \sqrt{\alpha_k} } \left( \bbx_k - \frac{\beta_k}{\sqrt{1 - \alphacumprod_k }}\, \bbepsilon_{\bbtheta^\star} \right).\label{eq:noisepred-mean-param-equivalence}
\end{align}
\noindent Thus $\bbepsilon_{\bbtheta^\star}$ and the fixed $\bbSigma_{\bbtheta} \!=\! \sigma_k^2 \bbI$ specify \eqref{eq:reverse-Gaussian-kernel} completely. For sampling, we construct a family of generally non-Markovian reverse processes~\cite{song2021denoising} that preserve the forward marginals \eqref{eq:forward-diffusion:conditionals},
\begin{align} \label{eq:ddim-sampler}
    \bbx_{k-1} = \sqrt{ \alphacumprod_{k-1} }\, \widehat{\bbx}_{0} + \sqrt{1 - \alphacumprod_{k-1} - \sigma_k^2}\, \bbepsilon_{\bbtheta^\star} + \sigma_k \bbw,
\end{align}
with $\bbw \sim \mathcal{N}(\bb0, \bbI)$. A single factor $\eta \in [0, 1]$ tunes the noise scale $\sigma_k$ [cf.~\eqref{eq:ddim-sigma}].
Each step draws $\bbx_{k-1}$ exactly from the reverse kernel~\eqref{eq:reverse-Gaussian-kernel} at $\eta = 1$, and smaller values yield near-deterministic and usually faster sampling.

At inference, we initialize $\bbx_K \sim \mathcal{N}(\bb0, \bbI)$ and iterate \eqref{eq:ddim-sampler} for $k = K, \ldots, 1$ to draw samples $\bbx_0 \sim \pgenstar(\cdot \cond \bbxi)$. These samples are distributed approximately according to the conditional data distribution $q_{\bbx \vert \bbxi}(\cdot \cond \bbxi)$ for the given conditioning $\bbxi = (\bbS, \bbu)$, thereby realizing the conditional generative model sought in \eqref{eq:generative-modeling}. To reduce the number of model evaluations, we run an accelerated variant of \eqref{eq:ddim-sampler} (see Appendix~\ref{app:accelerated-sampling}).

U-Nets have been instrumental in image generative models. Accordingly, we particularize $\bbepsilon_{\bbtheta}$ to a U-Graph Neural Network (U-GNN) that tailors U-Nets for graph signal denoising.


\section{Pooling \& Stride in Graph Neural Networks}
\label{sec:gnns}
We first review graph filters, convolutions, and a standard GNN layer. We then propose a GNN module with built-in pooling through selection matrices and stride, which Section~\ref{sec:ugnn-proposed} arranges into the U-GNN encoder--decoder pipeline.

\noindent\textbf{Graph filters and convolutions.}
Graph filters are the building blocks of GNNs and process input graph signals through successive graph shifts. A single shift $\bbS \bbx$ mixes each node's value with those of its one-hop neighbors, and $k$ repeated shifts yield $\bbS^k \bbx$, which aggregates at each node the signals within its $k$-hop neighborhood. A linear shift-invariant graph filter of order $K$ is the matrix polynomial $\bbh(\bbS) \coloneqq \sum_{k=0}^{K} h_k \bbS^k$ with taps (coefficients) $\bbh = [h_0, \ldots, h_K]$. It processes an input signal $\bbx$ by graph convolution,
\begin{align} \label{eq:graph-filter}
    \bby = \bbh *_{\bbS} \bbx = \bbh(\bbS)\, \bbx = \sum_{k=0}^{K} h_k \bbS^k \bbx,
\end{align}
and produces an output graph signal $\bby \in \reals^N$ as a weighted sum of the shifted signals $\bbS^k \bbx$.

\smallskip
\noindent\textbf{Graph neural networks.}
A GNN stacks graph convolutional layers, each composing a graph filter with a pointwise nonlinearity~\cite{gama2018convolutional}. The $\ell$th layer of an $L$-layer GNN reads
\begin{align} \label{eq:graph-convolution-layer}
    \bbX_{\ell} = \sigma_{\ell}\, \bbH_{\ell}(\bbS)\, \bbX_{\ell-1}
    = \sigma_{\ell}\!\left( \sum_{k=0}^{K} \bbS^k\, \bbX_{\ell-1}\,
    \bbTheta_{\ell,k} \right),
\end{align}
for $\ell = 1, \ldots, L$. Here, each scalar tap of~\eqref{eq:graph-filter} becomes a MIMO filter bank $\bbTheta_{\ell,k}$ that maps the $F_{\ell-1}$ input features in $\bbX_{\ell-1}$ to $F_{\ell}$ output features within $K$-hop neighborhoods, with learnable coefficients $\Theta_{\ell} = \{\bbTheta_{\ell,k}\}_{k=0}^{K}$ and a pointwise nonlinearity $\sigma_{\ell}$, e.g., a ReLU. Stacking these layers defines the $\Theta$-parametrized GNN $\bbPhi(\cdot, \bbS; \Theta) = \bbPhi_L(\cdot, \bbS; \Theta_L) \circ \cdots \circ \bbPhi_1(\cdot, \bbS; \Theta_1)$, with trainable parameters $\Theta = \{\Theta_{\ell}\}_{\ell=1}^{L}$. It maps an input $\bbX_0 \coloneqq \bbX \in \reals^{N \times F_0}$ to an output $\bbX_L \eqqcolon \bbY \in \reals^{N \times F_L}$,
\begin{align} \label{eq:gnn}
    \bbY = \bbPhi(\bbX, \bbS; \Theta).
\end{align}

The GNN in~\eqref{eq:gnn} is general-purpose but not tailored to diffusion-based generation. Since graph convolutions already extract features on arbitrary graphs, adapting the U-Net backbone additionally requires only node (un)pooling, which we introduce next together with a stride on the graph shift.

\smallskip
\noindent\textbf{Pooling.}
Consider a signal $\bbX \in \reals^{N \times F}$ on $N$ nodes. A sampling (selection) matrix $\bbD \in \{0,1\}^{N' \times N}$ with $N' < N$ retains the rows of $\bbI_N$ indexed by a sampling set $\Omega \subseteq \ccalV$ with $|\Omega| = N'$.\footnote{We present $\bbD$ as binary for exposition. We relax it to a soft selection mask during training and binarize it to pick $N'$ nodes at inference (Appendix~\ref{sec:ste}).} We learn $\Omega$, and hence $\bbD$, end-to-end, so it is input-dependent rather than fixed (Section~\ref{sec:node-selection}). We call $\rho \geq 1$ with $N' = \lfloor N / \rho \rfloor$ the \emph{down-sampling (pooling) factor} that sets the resolution reduction. By construction, $\bbD$ has orthonormal rows, $\bbD \bbD^\top = \bbI_{N'}$, and $\bbD^\top \bbD$ is a diagonal mask with ones on $\Omega$. The forward map $\bbZ = \bbD \bbX \in \reals^{N' \times F}$ down-samples $\bbX$. Its adjoint $\bbD^\top \bbZ$ up-samples back to $N$ nodes by \emph{zero-padding}, i.e., it restores the active entries on $\Omega$ and inserts zeros on the inactive nodes $\ccalV \setminus \Omega$.

A \emph{graph convolutional layer with pooling} processes a down-sampled signal $\bbZ_{\ell-1} \in \reals^{N' \times F_{\ell-1}}$ into $\bbZ_{\ell} \in \reals^{N' \times F_{\ell}}$,
\begin{align} \label{eq:pooled-graph-convolution-layer}
    \bbZ_{\ell} = \bbD\, \sigma_{\ell}\!\left( \sum_{k=0}^{K} \bbS^k\,
    \bbD^\top \bbZ_{\ell-1}\, \bbTheta_{\ell,k} \right).
\end{align}
\noindent We call~\eqref{eq:pooled-graph-convolution-layer} a \emph{lift--filter--reduce} layer. We lift $\bbZ_{\ell-1}$ to the full vertex set by zero-padding through $\bbD^\top$, filter on $\bbS$, and reduce the result back to the active set through $\bbD$. Because the inactive nodes are zero, each active node aggregates only the active values in its neighborhood on $\bbS$, which can be few under heavy pooling and motivates the stride introduced next.

\smallskip
\noindent\textbf{Stride.}
Modern CNNs, and U-Net variants in particular, often replace fixed pooling rules such as max- or sum-pooling with strided or dilated convolutions for learned down-sampling and receptive-field growth, e.g.,~\cite{nnunet2021isensee, yu2016multiscale}. A dilated convolution spaces its taps by a fixed step, so each tap reads from a point several positions away rather than from an immediate neighbor. On a graph, the natural spacing is the hop count. Thus, we introduce a stride (dilation) parameter $\gamma \in \mathbb{Z}^+$ and replace the unit shift $\bbS$ with the $\gamma$-hop shift $\bbS^\gamma$. The order-$K$ filter then aggregates from the dilated hops $\{0, \gamma, 2\gamma, \ldots, K\gamma\}$ rather than the consecutive hops $\{0, 1, \ldots, K\}$.
This yields the \emph{strided graph convolutional layer with pooling},
\begin{align} \label{eq:proposed-graph-convolution-layer}
    \bbZ_{\ell} = \bbPhi_{\ell}(\bbZ_{\ell-1}, \bbS^\gamma; \Theta_{\ell}, \bbD)
    = \bbD\, \sigma_{\ell}\!\left( \sum_{k=0}^{K} (\bbS^\gamma)^k\,
    \bbD^\top \bbZ_{\ell-1}\, \bbTheta_{\ell,k} \right),
\end{align}
which recovers the plain layer~\eqref{eq:graph-convolution-layer} when $\bbD = \bbI_N$ ($\rho = 1$) and $\gamma = 1$. The pooling factor $\rho$ and stride $\gamma$ serve complementary roles. Through $\bbD$, the factor $\rho$ sets how many nodes survive and hence the sparsity of the zero-padded signal. Through $\bbS^\gamma$, the stride $\gamma$ sets how far each tap reaches [cf. Remark~\ref{remark:stride}].

The strided GNN layer~\eqref{eq:proposed-graph-convolution-layer} keeps the lift--filter--reduce form of~\eqref{eq:pooled-graph-convolution-layer}, filtering on $\bbS^\gamma$ instead of $\bbS$. The computations remain on the original vertex set, and $\bbS$ is reused at every resolution without explicitly forming $\bbS^\gamma$ in our implementation. Instead, we filter through $\gamma K$ successive unit shifts and tap the signal every $\gamma$ hops. Thus, each layer reduces to sparse matrix--vector products, which we realize with the sparse routines of the \texttt{PyG} library. We refer to Appendix~\ref{app:strided-impl} for details.

\smallskip
\noindent\textbf{Reduced GSOs.}
The strided layer~\eqref{eq:proposed-graph-convolution-layer} also admits an equivalent form that operates directly on the $N'$-dimensional signals. Pushing $\bbD$ and $\bbD^\top$ through the sum defines the $k$-hop \emph{reduced graph shift operators (GSOs)},
\begin{align} \label{eq:reduced-GSOs}
    \bbS^{(k)} \coloneqq \bbD\, (\bbS^\gamma)^k\, \bbD^\top
    \in \reals^{N' \times N'},
\end{align}
whose one-hop instance $\bbS^{(1)}$ is the support of a coarse subgraph on $\Omega$. The family $\{\bbS^{(k)}\}_{k=0}^{K}$ then processes the low-dimensional signals directly,
\begin{align} \label{eq:proposed-graph-convolution-layer-reduced}
    \bbZ_{\ell} = \sigma_{\ell}\!\left( \sum_{k=0}^{K} \bbS^{(k)}\,
    \bbZ_{\ell-1}\, \bbTheta_{\ell,k} \right).
\end{align}
\noindent The two forms are equal because row selection commutes with the pointwise nonlinearity, $\bbD\, \sigma_{\ell}(\bbM) = \sigma_{\ell}(\bbD \bbM)$, and they differ only in the domain of computation. The reduced form avoids computation on the inactive nodes but is ill-suited to our setting. The reduced GSOs do not compose, i.e., $\bbS^{(m+n)} \neq \bbS^{(m)} \bbS^{(n)}$, since $\bbD^\top \bbD \neq \bbI_N$ in general. Therefore, each $\bbS^{(k)}$ must be formed separately rather than by powering $\bbS^{(1)}$. Moreover, $\bbD$ is input-dependent, so the family $\{\bbS^{(k)}\}$ is rebuilt on every forward pass, negating any gains from pre-computing. We therefore adopt the lift--filter--reduce form.

\smallskip
\noindent\textbf{GNNs with pooling and stride.}
Stacking $L$ strided graph convolutional layers with pooling [cf.~\eqref{eq:proposed-graph-convolution-layer}] defines the $(\Theta, \bbD)$-parametrized GNN module $\bbPhi(\cdot, \bbS; \Theta, \bbD) = \bbPhi_L(\cdot, \bbS^\gamma; \Theta_L, \bbD) \circ \cdots \circ \bbPhi_1(\cdot, \bbS^\gamma; \Theta_1, \bbD)$. It maps an input $\bbZ_0 \coloneqq \bbX \in \reals^{N' \times F_0}$ to an output $\bbZ_L \eqqcolon \bbY \in \reals^{N' \times F_L}$,
\begin{align} \label{eq:proposed-gnn}
    \bbY = \bbPhi(\bbX, \bbS; \Theta, \bbD),
\end{align}
with parameters $\Theta = \{\Theta_{\ell}\}_{\ell=1}^{L}$ and $\bbD \in \{0,1\}^{N' \times N}$. 
This GNN module is the main processing unit of each block in the U-GNN architecture, which is introduced next.

\begin{remark} \label{remark:stride}
In both views, the stride $\gamma$ offsets the sparsity introduced by down-sampling with $\bbD$. Under lift--filter--reduce, a stride such as $\gamma = \lfloor \sqrt{\rho}\, \rfloor$ keeps nodes from aggregating predominantly zeros. Under the reduced GSO, projecting through $\bbD$ may isolate nodes in the induced subgraph, and $\bbS^\gamma$ with $\gamma > 1$ restores connectivity. 
A suitable value depends on the average connectivity and the uniformity of $\bbD$.
\end{remark}

\begin{remark} \label{remark:reduced-filter}
A variant of~\eqref{eq:proposed-graph-convolution-layer-reduced} instead learns a genuine polynomial filter in the one-hop reduced GSO $\bbS^{(1)} = \bbD\, \bbS^\gamma\, \bbD^\top$, the principal submatrix of $\bbS^\gamma$ on $\Omega$. This is no longer equivalent to~\eqref{eq:proposed-graph-convolution-layer}, since sampling and filtering commute only at first order. Powering $\bbS^{(1)}$ confines every hop to $\Omega$, whereas the family $\bbS^{(k)}$ allows walks to traverse inactive nodes.
\end{remark}
\section{U-Graph Neural Networks}
\label{sec:ugnn-proposed}

%
%

\makeatletter\def\input@path{{figures/architecture/}}\makeatother

\begin{figure*}[ht!]\centering

  %
  \includegraphics[trim=11bp 168bp 11bp 9bp, clip, width=.95\linewidth]{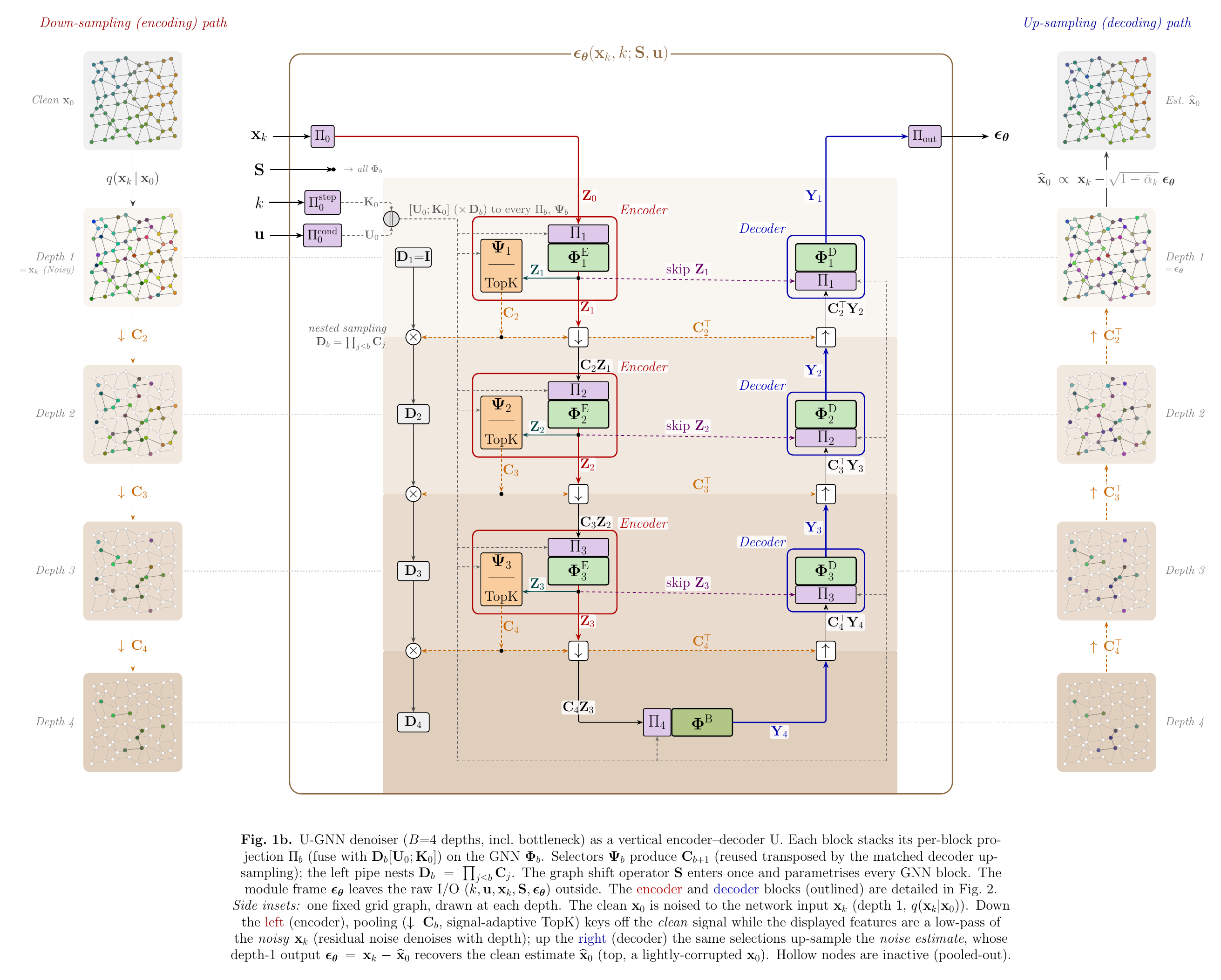}
\caption{\textbf{U-GNN denoiser architecture.}
A U-GNN of $B{=}4$ depths maps the noisy graph signal $\bbx_k$ to a noise estimate $\bbeps$. At each depth, a projection $\bbPi_b$ (purple) fuses the global embeddings $[\bbU_0;\bbK_0]$ of node states $\bbu$ and diffusion step $k$ into the signal path before a GNN module $\bbPhi_b^{\mathrm{E}/\mathrm{D}}$ (green), and decoder blocks additionally merge the encoder skips. Learned selectors $\bbPsi_b$ (orange) produce selection matrices $\bbC_{b+1}$ that down-sample the signal between depths and, transposed, up-sample it in the decoder. The shift operator $\bbS$ and the global embeddings $[\bbU_0;\bbK_0]$ are shared across all blocks.
Side panels trace one fixed input graph (signal) sample through all four depths, with the encoding (down-sampling) path on the left and decoding (up-sampling) on the right. On the left, a clean $\bbx_0$ (top-left) is corrupted into the input $\bbx_k$ at depth~$1$ via the forward diffusion process [cf.~\eqref{eq:forward-diffusion:conditionals} and \eqref{eq:reparametrization-trick}] and then down-sampled across depths by the selectors $\bbC_{b+1}$ ($\downarrow$). Node colors show a low-pass view of $\bbx_k$ whose residual noise fades with depth, and hollow markers denote inactive nodes. On the right, the transposes $\bbC_{b+1}^{\top}$ up-sample ($\uparrow$) the estimate to full resolution, yielding $\bbeps$ at depth~$1$, from which the clean $\widehat{\bbx}_0$ (top-right) is estimated via \eqref{eq:noisepred-denoiser-param-equivalence}.
}

  \label{fig:ugnn}
\end{figure*}

The GNN in~\eqref{eq:proposed-gnn} processes graph signals at a single node resolution fixed by the sampling matrix $\bbD$. To build the multi-resolution encoder--decoder hierarchy of a U-Net on top of this module, we first define the resolutions through depth-dependent selection matrices $\{\bbC_b\}_{b=1}^{B}$, where $\bbC_b \in \{0,1\}^{N_b \times N_{b-1}}$ selects $N_b$ nodes from the $N_{b-1}$ nodes active at the preceding depth, with $N_0 = N$ and $N = N_1 \geq N_2 \geq \cdots \geq N_B$. We form the composite (nested) sampling matrices,
\begin{align} \label{eq:sampling-matrix}
    \bbD_b \coloneqq \prod_{j=1}^{b} \bbC_j = \bbC_b \cdots \bbC_1
    \in \{0,1\}^{N_b \times N},
\end{align}
which relate the depth-$b$ nodes to the full $N$-node domain and induce a nested hierarchy of sampling sets $\Omega_{B} \subseteq \cdots \subseteq \Omega_1 \subseteq \ccalV$. We set $\bbD_1 = \bbC_1 = \bbI_N$ by convention, so the first depth operates at full resolution. Here, $\bbC_b$ and $\bbD_b$ carry the \emph{per-level} and \emph{cumulative} down-sampling factors $\rho_b \coloneqq N_{b-1}/N_b$ ($\rho_1 = 1$) and $\bar\rho_b \coloneqq \prod_{i=1}^{b} \rho_i = N/N_b$, respectively.
 
We call the resulting architecture a \emph{U-Graph Neural Network (U-GNN)}. It spans $B$ resolution levels $N = N_1 \geq \cdots \geq N_B$, with $B-1$ encoder--decoder block pairs and a bottleneck block at the coarsest resolution $N_B$. 
 
We present the U-GNN in two stages. We first fix a pure encoder--decoder cascade whose blocks are single encoding, decoding, and bottleneck GNN modules [cf.~\eqref{eq:proposed-gnn}], written $\enc_b$, $\dec_b$, or $\bneck_B$ with weights $\thetaenc_b$, $\thetadec_b$, and $\thetabneck_B$, respectively, along with the signal flow among them in~\eqref{eq:encoding-path}--\eqref{eq:bottleneck-block}. These equations include the skip connections, take the selection matrices $\bbC_b$ as given, and omit the node-state and diffusion-step embeddings. We then complete the blocks with projection (fusion) layers and, on the encoder side, node-selection heads (Sections~\ref{sec:conditioning} and~\ref{sec:node-selection}). These only enlarge the block interiors and parameter sets, leaving the composition here unchanged. The U-GNN overview in Fig.~\ref{fig:ugnn} and the complementary block interface details in Fig.~\ref{fig:ugnn-block} (deferred to Appendix~\ref{app:impl} due to limited space) may be consulted throughout this section.

The encoder blocks form an \emph{encoding path} $\enc = \enc_{B-1} \circ \cdots \circ \enc_1$, and the decoder blocks a \emph{decoding path} $\dec = \dec_1 \circ \cdots \circ \dec_{B-1}$. The encoding path extracts graph convolutional features at progressively coarser resolutions, while the decoding path reverses this progression by restoring resolution and reintroducing fine details through skip connections.
 
We describe each path in turn, starting from the input features $\sigenc_0$. Each encoder block maps its input to an encoded signal $\sigenc_b \in \reals^{N_b \times F_b}$,
\begin{align}\label{eq:encoding-path}
    \sigenc_b = \enc_b \left( \bbC_b \sigenc_{b-1},\, \bbS;\, \thetaenc_b,\, \bbD_b
    \right),
\end{align}
for $b = 1, \ldots, B-1$, where $\bbC_b$ down-samples the depth-$(b{-}1)$ feature onto the $N_b$ nodes processed at depth $b$. At the top level, $\bbC_1 = \bbI_N$ and $\enc_1$ operates at full resolution $N_1 = N$.
 
Each decoder block receives the coarser decoded signal $\sigdec_{b+1} \in \reals^{N_{b+1} \times F_{b+1}}$ from depth $b+1$, up-samples it onto the $N_b$-node support through $\bbC^\top_{b+1}$, and combines it with the matched encoder feature $\sigenc_b$ through a skip connection,
\begin{align}\label{eq:decoding-path}
    \sigdec_b = \dec_b \left( \big[ \bbC^\top_{b+1} \sigdec_{b+1};\, \sigenc_b \big],\,
    \bbS;\, \thetadec_b,\, \bbD_b \right),
\end{align}
for $b = B-1, \ldots, 1$, where $[\,\cdot\,;\,\cdot\,]$ denotes the skip combination. The decoded signal $\sigdec_b$ then passes to the next decoder block or to the read-out layer at $b = 1$.
 
The two paths connect at depth $B$ through the bottleneck block, which acts as a $B$th encoder block,
\begin{align} \label{eq:bottleneck-block}
    \sigdec_{B} = \bneck_B \left( \bbC_{B} \sigenc_{B-1},\, \bbS;\, \thetabneck_B,\,
    \bbD_{B} \right).
\end{align}
The bottleneck output $\sigdec_{B}$ enters the decoding path~\eqref{eq:decoding-path} at depth $b = B-1$, and the final decoder output $\sigdec_1 \in \reals^{N \times F_0}$ is mapped by a read-out layer to the model target, i.e., the noise prediction $\bbepsilon_{\bbtheta}$ here. 
We note that all blocks share the same $\bbS$, and a depth-dependent stride $\gamma_b$ is implicit in the notation [cf.~\eqref{eq:proposed-graph-convolution-layer}]. Appendix~\ref{app:strided-impl} specifies how $\gamma_b$ is set and how each block evaluates its strided convolutions on the shared $\bbS$. 

\subsection{Input Processing and Block-Level Conditioning}
\label{sec:conditioning}

The U-GNN realizes the denoiser $\bbepsilon_{\bbtheta}(\bbx_k, k; \bbxi)$ with conditioning $\bbxi = (\bbS, \bbu)$. It incorporates the conditioning along two routes. The graph enters as the shared shift operator $\bbS$ of every GNN module [cf.~Section~\ref{sec:gnns}], while the node states $\bbu$ enter as side information, embedded and fused into each block together with an embedding of the diffusion step $k$. Note that the step $k$ is a model input and not part of the conditioning $\bbxi$. We now detail the embedding pipeline. 

A read-in layer $\projlayer_0$ maps the noisy signal $\bbx_k \in \reals^{N}$ to the initial node embeddings $\sigenc_0 \in \reals^{N \times F_0}$ that enter the first encoder block. The node states $\bbu \in \reals^{N \times U}$ and the step $k$ are embedded separately into global representations shared across all blocks. A node-wise MLP $\projlayercond_0$ (a shared per-node MLP with no cross-node mixing) maps $\bbu$ to per-node embeddings $\bbU_0 = \projlayercond_0(\bbu) \in \reals^{N \times F_0}$. A diffusion-step encoder $\projlayerdiffusionstep_0$, a sinusoidal positional encoding followed by a node-wise MLP, maps the scalar $k$ to an $F_0$-dimensional vector broadcast to all nodes as $\bbK_0 = \bbone_N\, \projlayerdiffusionstep_0(k)^\top \in \reals^{N \times F_0}$.
 
At each depth $b$, a fusion layer $\projlayer_b$ merges the main signal path with these global embeddings before the GNN module. The block input $\bbV_b$ and the down-sampled embeddings $\bbD_b\,[\bbU_0;\, \bbK_0]$ are each projected to a common width, merged along the feature dimension, and mapped to $F_b$ channels,
\begin{align} \label{eq:block-input-fusion}
    \bbP_b = \projlayer_b\big(\bbV_b,\; \bbD_b\, [\bbU_0;\, \bbK_0]\big).
\end{align}
The factor $\bbD_b$ restricts the global embeddings to the $N_b$ nodes active at depth $b$, matching the support of $\bbV_b$, and the per-input projection reconciles their differing feature (channel) widths. The block input depends on the path. For $b$th encoder block, $\bbV_b = \bbC_b\, \sigenc_{b-1}$ down-samples the depth-$(b{-}1)$ feature onto the $N_b$ active nodes. For $b$th decoder block, the coarser output $\sigdec_{b+1} \in \reals^{N_{b+1} \times F_{b+1}}$ is up-sampled through $\bbC_{b+1}^\top$, concatenated with the encoder skip $\sigenc_b$ along the feature dimension, and mapped to $F_b$ channels by a learned skip-projection $\projlayerskip_b: \reals^{F_{b+1} + F_b} \to \reals^{F_b}$,
\begin{align} \label{eq:skip-fusion}
    \bbV_b = \projlayerskip_b \left( \big[ \bbC_{b+1}^\top \sigdec_{b+1};\, \sigenc_b
    \big] \right).
\end{align}
The fusion layer $\projlayer_b$ is shared between the matched encoder and decoder blocks, so its weights appear in both $\thetaenc_b$ and $\thetadec_b$, while the GNN module and skip-projection remain block-specific. The node-state and diffusion-step embeddings enter each block by input substitution. In~\eqref{eq:encoding-path} and~\eqref{eq:bottleneck-block}, the module input $\bbC_b\, \sigenc_{b-1}$ is replaced by the fused $\bbP_b$. In~\eqref{eq:decoding-path}, the skip combination $[\bbC^\top_{b+1}\sigdec_{b+1};\, \sigenc_b]$ is first mapped by $\projlayerskip_b$ [cf.~\eqref{eq:skip-fusion}] and then fused into $\bbP_b$. The GNN modules and the block composition remain unchanged, and only the module inputs differ. Figure~\ref{fig:ugnn} shows this separation, with the fusion drawn as the purple projection $\projlayer_b$ preceding each green GNN module, and Fig.~\ref{fig:ugnn-block} (Appendix~\ref{app:impl}) details the block interface. Finally, a read-out layer $\projlayer_{\mathrm{out}}$, a node-wise MLP, maps the final decoder output $\sigdec_1 \in \reals^{N \times F_0}$ to the prediction $\bbepsilon_{\bbtheta} \in \reals^{N}$.

\subsection{Down-sampling (Pooling) via Learned Node Selection}
\label{sec:node-selection}

The selection matrices $\bbC_b$ and their compositions $\bbD_b$ determine the active nodes at each depth. Rather than fixing them in advance, we learn them end-to-end by letting each encoder block score its own GNN output and select the active nodes propagated to the next depth. The node-selection head of encoder block $b$ derives $\bbC_{b+1}$ from the encoded feature $\sigenc_b$, for $b = 1, \ldots, B-1$, producing $\bbC_2, \ldots, \bbC_B$. The top level performs no selection since $\bbC_1 = \bbD_1 = \bbI_N$, and the deepest selection $\bbC_B$ down-samples into the bottleneck, which itself does not further down-sample.
 
We describe the selection operation using two index systems. \emph{Global} indices label nodes in $\ccalV$, with the active set $\Omega_b \subseteq \ccalV$, $|\Omega_b| = N_b$, nested as in the beginning of Section~\ref{sec:ugnn-proposed}. \emph{Local} indices $[N_b] \coloneqq \{1, \ldots, N_b\}$ enumerate only the active nodes in a fixed order. The composite matrix $\bbD_b \in \{0,1\}^{N_b \times N}$ maps a global-indexed signal to its local form, and $\bbD_b^\top$ scatters a local-indexed signal back to the $N$-node domain. The selection below acts in the local space $[N_b]$. 
 
Each encoder block is equipped with a head that scores the active nodes and keeps the highest-scoring ones. Concretely, a node-wise MLP $\bbPsi_b$, with parameters $\thetasel_b$, takes the block output $\sigenc_b$, along with the node and step embeddings restricted to the active nodes, and returns a per-node score vector,
\begin{align} \label{eq:node-selector-block}
    \bbv_b = \bbPsi_b \left( \sigenc_b,\, \bbD_b\,[\bbU_0;\, \bbK_0];\, \thetasel_b
    \right) \in \reals^{N_b}.
\end{align}
The dependence on $\bbD_b\,[\bbU_0;\, \bbK_0]$, as in~\eqref{eq:block-input-fusion}, allows the selection to adapt to the node states and the diffusion step. Given a down-sampling factor $\rho_{b+1} > 1$, we keep the $N_{b+1} = \lfloor N_b / \rho_{b+1} \rfloor$ highest-scoring nodes,
\begin{align} \label{eq:topk}
    \mathcal{S}_b = \mathrm{TopK}(\bbv_b,\, N_{b+1}) \subseteq [N_b].
\end{align}
The selection matrix $\bbC_{b+1} = \big[ \bbI_{N_b} \big]_{\mathcal{S}_b,\,:} \in \{0,1\}^{N_{b+1} \times N_b}$ keeps the rows of $\bbI_{N_b}$ indexed by $\mathcal{S}_b$, and the composite matrix updates as $\bbD_{b+1} = \bbC_{b+1} \bbD_b$ [cf.~\eqref{eq:sampling-matrix}]. The retained global set $\Omega_{b+1} \subseteq \Omega_b$ collects the nonzero columns of $\bbD_{b+1}$, so $\mathcal{S}_b$ and $\Omega_{b+1}$ express the same selection in local and global coordinates, respectively. 

The decoder performs no selection. At each depth, the encoder-produced $\bbC_{b+1}$ is reused by the matched decoder for up-sampling through $\bbC_{b+1}^\top$ [cf.~\eqref{eq:skip-fusion}], while $\bbD_b$ parametrizes the depth-$b$ GNN modules [cf.~\eqref{eq:encoding-path}--\eqref{eq:bottleneck-block}]. 

The Top-K operation in~\eqref{eq:topk} is non-differentiable in $\bbv_b$. Therefore, we
train the head $\bbPsi_b$ with a straight-through estimator (STE)~\cite{bengio2013estimating} that keeps the hard selection $\mathcal{S}_b$ in the forward pass, while gradients reach the
scores $\bbv_b$ through a differentiable sigmoid surrogate of the binary selection
mask. 
Appendix~\ref{sec:ste} details the surrogate mask, and Appendix~\ref{app:sec:pooling} integrates classical fixed-pooling rules with the U-GNN.

\section{Numerical Results}
\label{sec:experiments}

We evaluate the proposed U-GNN denoiser on stock price forecasting for the S\&P~500 index and wireless resource allocation (WRA) in a power control setup. Both use the same depth-$B{=}4$ U-GNN backbone $\bbepsilon_{\bbtheta}$, diffusion noise schedules, and optimizers. They differ only in the signal and conditioning interface that each task dictates---a
multi-step return trajectory for forecasting versus a single power allocation for resource allocation. Table~\ref{tab:impl} summarizes the model, diffusion, and optimization hyperparameters. We detail the shared and application-specific configurations in Appendix~\ref{app:impl}.


\def\baseCh{64}            
\def\numPoolLevels{3}      
\def\poolFactor{2}         
\def\gnnLayers{2}          
\def\gnnHops{2}            
\def\maxStride{2}          
\def\numBottleneck{2}      
\def\dropoutRate{0.1}
\def\condEmbDim{128}       
\def\timeEmbDim{128}       
\def\signalCh{1}           

\def\numDiffSteps{500}     
\def\betaStart{10^{-4}}\def\betaEnd{2\times10^{-2}}
\def\numSampleSteps{100}   
\def\ddimEta{0.2}
\def\numTraj{100}

\def\learnRate{10^{-4}}
\def\weightDecay{10^{-4}}
\def\adamBetas{(0.9,0.98)}
\def\gradClip{1.0}
\def\maxEpochs{5000}
\def\warmupFrac{1\%}       
\def\minLrFrac{5\%}        
\def\decayFrac{0.8}        

\def\numParamsStock{1.44}  
\def\batchStock{64}
\def\covLevels{\{80,90,95,99\}}  


\def\numParamsWra{0.63}    
\def\batchWra{800}
\def\condChannelsWra{2}    

\def\numStocks{468}            
\def\numUniverse{500}          
\def\numSectors{11}            
\def\numTradingDays{2353}      
\def\dataStart{2016-09-06}
\def\dataEnd{2026-02-13}
\def\dataYears{9.4}
\def\numWindows{2328}          
\def\Thval{20}                 
\def\Tpval{5}                  
\def\numFeat{12}               
\def\numFund{15}               
\def\corrMethodStock{Spearman}
\def\corrThresholdStock{0.7}
\def\sectorBonusStock{0.05}
\def\minCoverageStock{0.95}
\def\numEdgesStock{5092}       
\def\edgeDensityStock{4.7}     
\def\avgDegreeStock{22}
\def\maxDegreeStock{51}
\def\numSplitChunks{10}        
\def\rsiWin{14}                
\def\macdShort{12}\def\macdLong{26}   

\def\alrW{\{5,10,21,42\}}      

\def\revinBlend{0.7}           
\def\revinAlpha{1.107}         

\def\grwShrinkage{10}      

\def \deltaTms {10}              
\def \numFadingSteps {500}       
\def \ergodicWindowSteps {5}     
\def \slotDurationMs {50}        
\def \numTimeSlots {100}         
\def \bandwidthInMHz {40}
\def \noisePSDIndBmPerHz {-174}
\def \carrierFreq {2.4}
\def \PmaxInMW {10}              
\def \PmaxIndBm {10}
\def \numNodes {400}
\def \numRx {\numNodes}
\def \rmin {0.6}
\def \fmin {0.6}
\def \minTxRxSep {20}            
\def \maxTxRxSep {100}           
\def \minTxTxSep {50}            
\def \shadowingStd {7}           
\def \plExpShort {2}             
\def \plExpLong {4}              
\def \plBreakpoint {100}         
\def \numFadingPaths {100}       
\def \mobilitySpeed {1}          
\def \numNetsPerDensity {32}
\def \numDensities {4}
\def \topK {10}
\def \numExpertSamples {200}     
\def \pdHistoryWindow {1000}     
\def \pdDualStep {0.2}           
\def \pdHidden {64}              
\def \pdLayers {3}               
\def \pdHops {2}                 
\def\numTrainSamples{16000}   
\def \numTotalSamples {25600}    

\def \h {\bbH}
\def \x {\bbx}

\def \deploymentRultralow {7800}
\def \deploymentRlow {7000}
\def \deploymentRmid {6300}
\def \deploymentRhigh {5800}

\pgfmathsetmacro{\numNets}  { \numNetsPerDensity * 4 }
\pgfmathsetmacro{\deploymentAreaultralow} {(\deploymentRultralow / 1000) * (\deploymentRultralow / 1000)}
\pgfmathsetmacro{\deploymentArealow}      {(\deploymentRlow / 1000) * (\deploymentRlow / 1000)}
\pgfmathsetmacro{\deploymentAreamid}      {(\deploymentRmid / 1000) * (\deploymentRmid / 1000)}
\pgfmathsetmacro{\deploymentAreahigh}     {(\deploymentRhigh / 1000) * (\deploymentRhigh / 1000)}
\pgfmathsetmacro{\densityultralow} { \numNodes * (1 / \deploymentAreaultralow)}  
\pgfmathsetmacro{\densitylow}      { \numNodes * (1 / \deploymentArealow)}       
\pgfmathsetmacro{\densitymid}      { \numNodes * (1 / \deploymentAreamid)}       
\pgfmathsetmacro{\densityhigh}     { \numNodes * (1 / \deploymentAreahigh)}      

\pgfmathsetmacro{\numTestNets} {\numNets * 2 / 8}
\pgfmathsetmacro{\numTestRx}   {\numTestNets * \numRx}

\begin{table}[t]
  \centering
  \caption{Summary of U-GNN configuration.}
  \label{tab:impl}
  \begin{tabular}{@{}lcc@{}}
    \toprule
    \textbf{Hyperparameter} & \textbf{S\&P~500} & \textbf{WRA} \\
    \midrule
    \multicolumn{3}{@{}l}{\textit{U-GNN (depth $B=4$)}}\\
    Pooling levels (factor)       & \multicolumn{2}{c} {$\numPoolLevels$ ($\rho{=}\poolFactor$, $\bar{\rho}_B{=}8$) } \\
    Channel widths $F_0 = \cdots = F_B$              & \multicolumn{2}{c}{\baseCh} \\
    GNN layers $L$ / hops $K_\ell$            & \multicolumn{2}{c}{$\gnnLayers$ / $\gnnHops$} \\
    GNN stride $\gamma_b$   & \multicolumn{2}{c}{$\min \left \{ \left \lfloor \bar{\rho}_b \right \rfloor, \gamma_{\max} = 2 \right \}$} \\
    Embedding dim (step, cond.)   & \multicolumn{2}{c}{\timeEmbDim} \\
    Dropout                       & \multicolumn{2}{c}{\dropoutRate} \\
    Conditioning channels $U$     & \numFeat & \condChannelsWra \\
    Conditioning fusion           & cross-attn.\ & concat. \\
    \midrule
    \multicolumn{3}{@{}l}{\textit{Diffusion}}\\
    Diffusion noise steps $K$                & \multicolumn{2}{c}{\numDiffSteps} \\
    Noise schedule $\beta$               & \multicolumn{2}{c}{\hspace{-5em}linear $\left( \beta_1 = \betaStart, \beta_{\numDiffSteps} = \betaEnd \right)$} \\
    DDIM steps ($\eta$)           & \multicolumn{2}{c}{$\numSampleSteps$ ($\ddimEta$)} \\
    \midrule
    \multicolumn{3}{@{}l}{\textit{Optimization}}\\
    Optimizer                     & \multicolumn{2}{c}{AdamW, $\beta = \adamBetas$} \\
    Learning rate schedule  & \multicolumn{2}{c}{Cosine anneal ($\learnRate$)} \\
    Epochs                        & \multicolumn{2}{c}{\maxEpochs} \\
    Batch size                    & \batchStock & \batchWra \\
    \bottomrule
  \end{tabular}
\end{table}

\subsection{Stock Price Forecasting}
\label{sec:spatio-temporal-experiments}

We cast stock price forecasting as a spatio-temporal generative task. Given recent market history for a panel of stocks, the goal is to sample from the conditional distribution of their short-horizon future log-return trajectories.

\begin{table*}[t]
  \centering
  \caption{\textbf{Forecast accuracy and distributional fidelity comparisons on the S\&P~500 test split.} Best values per column are shown in \textbf{bold}. The \textbf{U-GNN} outperforms the GRW baseline in almost every accuracy, calibration and stylized-fact metrics.}
  \label{tab:sp500-returns}
  \setlength{\tabcolsep}{3pt}
  \begin{tabular*}{\textwidth}{@{\extracolsep{\fill}} l cccc cccc c ccc @{}}
    \toprule
    & \multicolumn{4}{c}{Returns} & \multicolumn{4}{c}{Price (\$)}
    & Direction & \multicolumn{3}{c}{Stylized-fact gaps} \\
    \cmidrule(lr){2-5}\cmidrule(lr){6-9}\cmidrule(lr){10-10}\cmidrule(lr){11-13}
    Model
    & CRPS\,$\downarrow$ & MIS$_{90}$\,$\downarrow$ & RMSE\,$\downarrow$ & MAE\,$\downarrow$
    & CRPS\,$\downarrow$ & MIS$_{90}$\,$\downarrow$ & RMSE\,$\downarrow$ & MAE\,$\downarrow$
    & MV-DA\,$\uparrow$
    & Vol\,$\downarrow$ & Mom\,$\downarrow$ & Kurt\,$\downarrow$ \\
    \midrule
    \textbf{U-GNN} & \textbf{1.01} & \textbf{9.28} & \textbf{1.96} & \textbf{1.36}
          & \textbf{2.69} & \textbf{25.17} & \textbf{10.03} & \textbf{3.62}
          & \textbf{0.53}
          & \textbf{0.06} & \textbf{0.03} & 12.96 \\
    GRW   & 1.08 & 9.69 & 2.16 & 1.42
          & 3.43 & 30.01 & 13.96 & 4.61
          & 0.49
          & 0.10 & 0.05 & \textbf{7.71} \\
    \bottomrule
  \end{tabular*}
  \par\vspace{3pt}
  \begin{minipage}{\textwidth}
    \footnotesize
    \textbf{MV-DA}: cumulative-horizon majority-vote direction accuracy.
    \textbf{Vol}, \textbf{Mom}, \textbf{Kurt}: absolute gap
    $|s_{\text{gen}}-s_{\text{real}}|$ between generated and real
    volatility-clustering, momentum, and excess-kurtosis statistics.
    Continuous ranked probability score (CRPS), mean  90\% prediction interval score (MIS$_{90}$), root mean squared error (RMSE), and mean absolute error (MAE) are reported on both the return and nominal price scales.
  \end{minipage}
\end{table*}

\begin{figure*}[t!]
    \centering
    \includegraphics[width=.328\linewidth]{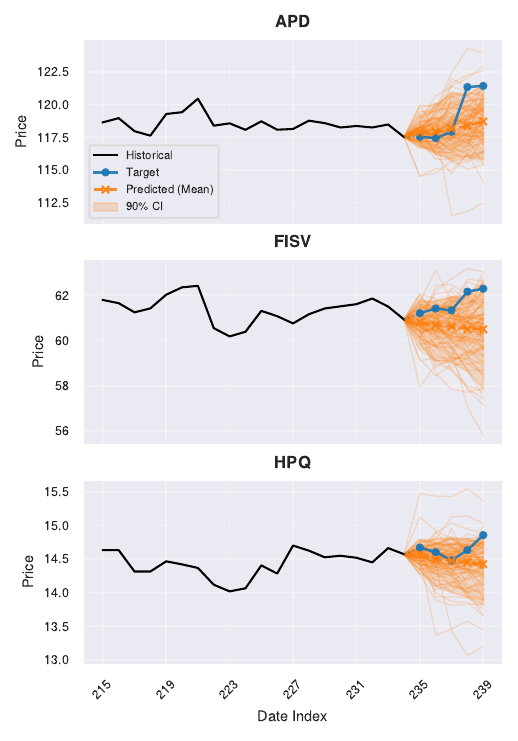} 
    \hfill 
    \includegraphics[width=.328\linewidth]{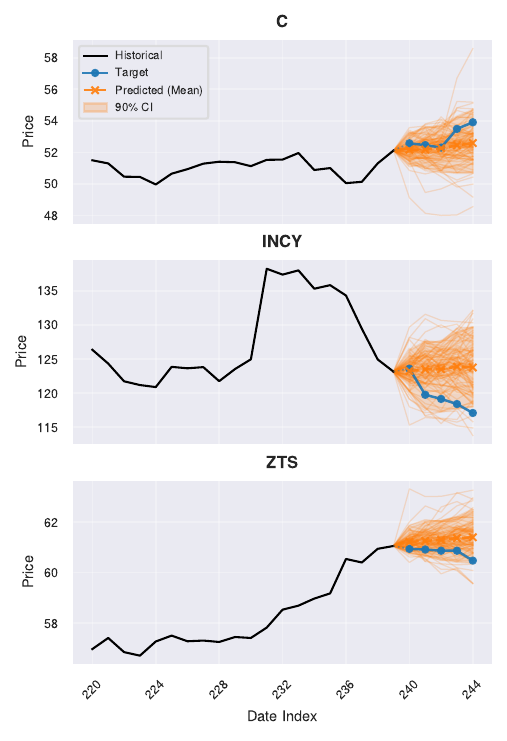} 
    \hfill 
    \includegraphics[width=.328\linewidth]{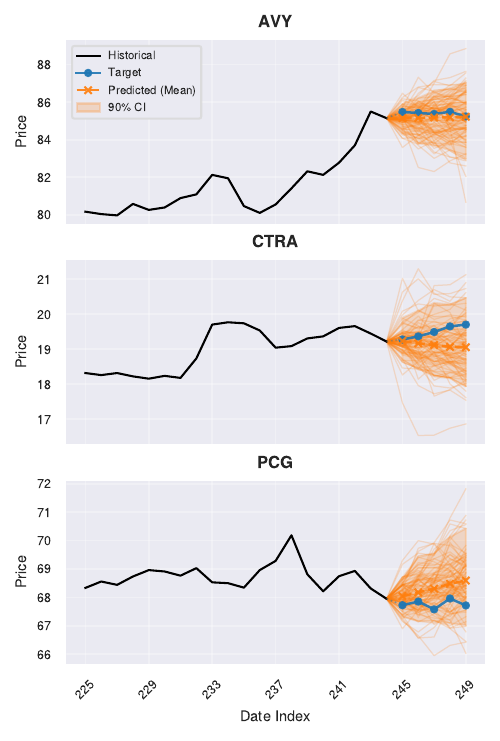}
    \caption{\textbf{Example S\&P~500 forecasting trajectories from U-GNN}. Each column is a different test window with three panels for arbitrarily chosen stocks. In each panel, the solid black line is the observed history whereas the solid blue line is the ground-truth future, the thin orange lines are individual trajectories sampled from the diffusion ensemble, and the solid orange line is their mean, over the forecast horizon. The spread of the sampled trajectories reflects the forecast uncertainty.}
    \label{fig:stock_prices}
\end{figure*}
\begin{figure*}[t!]
  \centering
  \begin{minipage}[t]{0.33\textwidth}
    \centering
 \begin{subfigure}{\linewidth}
    \centering
    \captionsetup{margin={0.154\linewidth,0pt}}%
    \vspace*{5pt}
    \includegraphics[trim=0bp 0bp 249.13bp 0bp, clip, width=\linewidth]{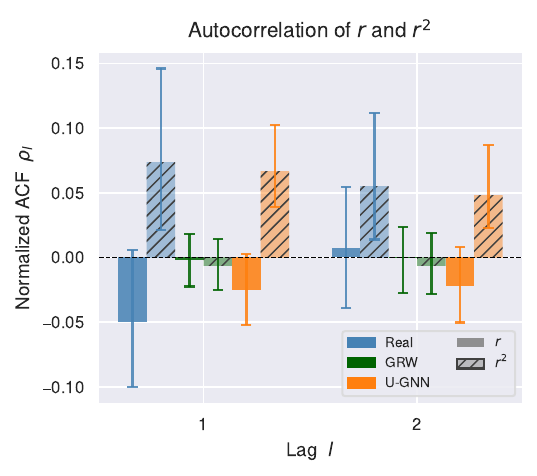}%
    \par\vspace*{-5pt}
    \caption{}\label{fig:acf}
  \end{subfigure}
    \\[1pt]
    \begin{subfigure}{\linewidth}
      \centering
      \captionsetup{margin={0.126\linewidth,0pt}}
      \includegraphics[trim=257.6bp 0bp 0bp 0bp, clip, width=\linewidth]{figures/sp500/fig2_structural_comparison.pdf}
      \par\vspace*{-5pt}
      \caption{
      }
      \label{fig:eig}
    \end{subfigure}
  \end{minipage}\hfill
  \begin{minipage}[t]{0.63\textwidth}
    \centering
    \begin{subfigure}{\linewidth}
      \centering
      \includegraphics[trim=0bp 234.44bp 0bp 0bp, clip, width=\linewidth]{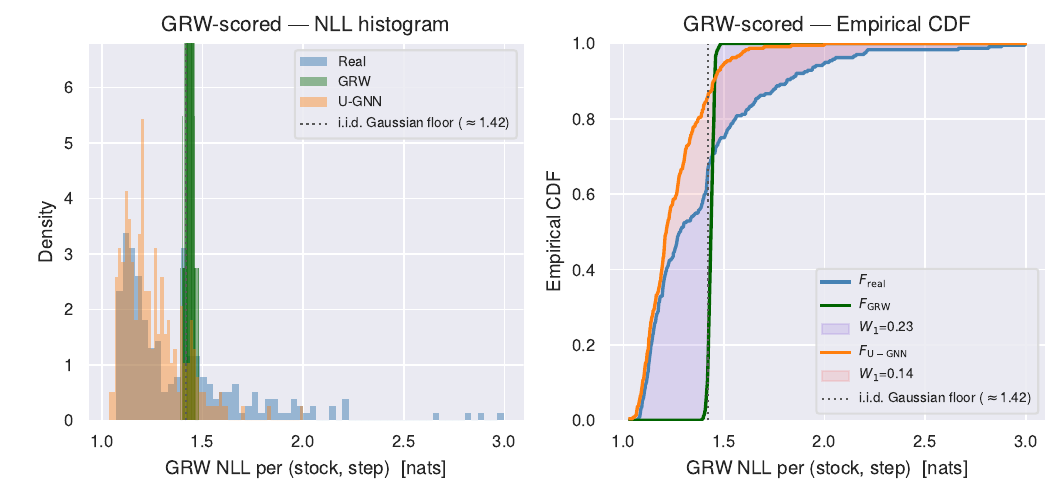}
      \caption{
      }
      \label{fig:grw-nll}
    \end{subfigure}\\[1pt]
    \begin{subfigure}{\linewidth}
      \centering
      \includegraphics[trim=0bp 0bp 0bp 232.5bp, clip, width=\linewidth]{figures/sp500/fig3_nll_comparison.pdf}
      \caption{
      }
      \label{fig:ugnn-nll}
    \end{subfigure}
  \end{minipage}
  \caption{\textbf{Distributional fidelity diagnostics on the S\&P~500 test split.} \textbf{(a)} Temporal structure: autocorrelation of returns $r$ and squared returns $r^2$ for real and generated series, probing linear predictability and volatility clustering. \textbf{(b)} Spatial (cross-sectional) structure: the top-5 eigenvalues of the return covariance matrix $\hat{C}$, capturing how well each model reproduces inter-stock correlation. \textbf{(c)--(d)} Histograms and empirical CDFs of per-window negative log-likelihood (NLL): values are computed under the GRW model's analytic likelihood \textbf{(c)} and as ELBO estimates under the trained U-GNN \textbf{(d)}, each evaluated on the real data and on samples drawn from both models. Close overlap between the generated and real distributions indicates higher distributional realism; U-GNN's samples track the real statistics more closely than the GRW baseline across all four panels.}
  \label{fig:distributional-realism}
\end{figure*}

\smallskip
\noindent \textbf{Setup.} We use daily S\&P~500 data obtained with the open-source \texttt{yfinance} library. After aligning all stocks on a common trading day axis and dropping those with insufficient coverage, we retain $N=\numStocks$ stocks across $\numSectors$ GICS sectors over $\numTradingDays$ trading days (\dataStart\ to \dataEnd). Let $\rday{t}\in\reals^{N}$ collect the daily log returns on trading day $t$, where $[\rday{t}]_i$ is the log ratio of stock $i$'s consecutive closing prices. For each stock and day, we gather $U=\numFeat$ market features into $\uday{t}\in\reals^{N\times U}$, comprising the open, high, low, and standardized closing prices; the log return $\rday{t}$ and its trailing moving averages; the log-volume; and two technical indicators, RSI and MACD. 

We build a static, undirected graph $\ccalG$ from long-term company fundamentals (e.g., market capitalization and price-to-earnings ratios). Its adjacency $\bbW$ combines the rank correlation between two stocks' fundamental profiles with a same-sector bonus. We threshold and spectrally normalize $\bbW$ to obtain the GSO $\bbS$, which is shared across all forecast windows. 

We form data samples with a sliding window of length $(\Th+\Tp)$ over the panel. In each window, the first $\Th$ days are the conditioning history, stacked into $\bbu=[\,\uday{t-\Th+1},\ \cdots,\ \uday{t}\,]\in\reals^{N\times\Th\times U}$, and the next $\Tp$ days are the forecast target, stacked into the clean graph signal $\bbx_0=[\,\rday{t+1},\ \cdots,\ \rday{t+\Tp}\,]\in\reals^{N\times\Tp}$. We reversibly instance-normalize (RevIN) each target window \cite{kim2022reversible}, using statistics from its conditioning block alone,
and we invert this normalization on the DDIM-generated samples before scoring. The U-GNN treats the $\Tp$ horizon steps as a temporal axis. It applies graph convolutions on $\bbS$ at each step, couples the steps through interleaved temporal-convolution layers, and fuses a per-node summary of the $\Th$ history steps into the conditioning context by cross-attention.

We set $\Th=\Thval$ and $\Tp=\Tpval$, yielding $\numWindows$ windows in total. We partition them using an interleaved chronological scheme. We split the trading-day axis into $\numSplitChunks$ equal chunks. Within each chunk, we assign the earliest $80\%$ days to training, the next $10\%$ to validation, and the final $10\%$ to testing, then pool the matching parts across chunks. This ordering keeps each test block later than its training block, so the model is never fitted on days that follow its test period. Interleaving also spreads anomalous regimes, such as COVID-19, across all three splits rather than concentrating them in one, which mitigates the test distribution shift that a single contiguous split would create.


\smallskip
\noindent \textbf{Results.} We assess the U-GNN forecasts qualitatively through the sample trajectories of Fig.~\ref{fig:stock_prices} and quantitatively through the accuracy and distributional fidelity metrics of Table~\ref{tab:sp500-returns}, both computed on the held-out test split with $\numTraj$ sampled trajectories per window. We benchmark U-GNN against a geometric random walk (GRW), in which each stock's future log-returns are drawn i.i.d.\ from a fixed per-stock Gaussian. Its drift and volatility are estimated once on the training set, with 
shrinkage toward the market-wide statistics, and they are held fixed at inference, independently of the conditioning window.

U-GNN outperforms the GRW on nearly every metric, at both the return and price scales. It reduces price CRPS, RMSE, and MAE by $22\%$, $28\%$, and $21\%$, respectively, and recovers directional information unavailable to a random walk, raising majority-vote direction accuracy above chance ($0.53$ vs.\ $0.49$). Beyond point and interval accuracy, U-GNN reproduces stylized facts of returns. It substantially shrinks the volatility-clustering gap ($0.06$ vs.\ $0.10$) and reduces the momentum gap. The one exception is excess kurtosis, where it overshoots the empirical heavy tails. This overshoot is the artifact of a desirable property. The GRW confines nearly all of its mass to a band around the mean, whereas U-GNN places non-negligible probability on the rare, far-from-mean moves that real markets exhibit. At the reported checkpoint it allocates somewhat more mass to these extremes than the data warrant, and we read this gap as a fidelity--sharpness trade-off rather than a fixed limitation. Checkpoints with a near-vanishing kurtosis gap are attainable, but only by sacrificing sharpness. Sharp forecasts emerge early in training and distributional fidelity improves later, with the reported model sitting toward the sharper end (Fig.~\ref{fig:ugnn-nll}).
 
These dynamics are evident in the individual forecasts of Fig.~\ref{fig:stock_prices}. The sampled trajectories spread out from the last observed price, and the spread grows over the forecast days, while their mean stays close to the realized path. Each sample tends to track the recent behavior of its stock, picking up its trend and volatility rather than flattening into a straight line or spreading evenly in all directions. A few paths break from the bundle to trace larger excursions, which is the generative model anticipating the occasional market swings noted above.

Fig.~\ref{fig:distributional-realism} assesses the distributional fidelity of U-GNN samples against those of GRW along four axes. Fig.~\ref{fig:acf} shows that U-GNN matches the autocorrelation signature of
stock returns, which is near-zero for $r$ and persistent for $r^2$, with the latter implying strong volatility clustering. Fig.~\ref{fig:eig}
pertains to the cross-sectional structure, where U-GNN tracks the leading eigenvalues of the return covariance $\hat{C}$ that capture the dominant market directions, whereas the GRW baseline flattens them. The remaining two plots take a score-based view. Each assigns every evaluation window a scalar score, then compares the resulting distribution to the real one through the $1$-Wasserstein ($W_1$) distance between their empirical CDFs. They differ only
in which model supplies the score.

Fig.~\ref{fig:grw-nll} uses the GRW score, the analytic NLL per (stock, step) in nats, for which an i.i.d.\ Gaussian is lower-bounded by its differential entropy $\tfrac{1}{2}\ln(2\pi e)\approx 1.42$ (dotted line). Scored by the homoskedastic Gaussian that generated them, the GRW samples collapse onto a near-degenerate spike essentially at this floor, so every window is equally surprising, and the score carries no volatility information. The real returns, instead, spread out with a rightward skew, as calm windows fall below the floor and heavy-tailed ones, beyond the reach of the GRW, are pushed into the tail. U-GNN's samples sit just left of the floor (mean $\approx 1.25$), which the GRW scorer even rates as more typical than an i.i.d.\ draw. Even under the baseline's own likelihood, U-GNN lands closer to the real distribution ($W_1{=}0.14$) than GRW's own samples do ($W_1{=}0.23$).

Fig.~\ref{fig:ugnn-nll} instead scores each window by a uniformly weighted version of U-GNN's training loss in~\eqref{eq:ddpm-noise-training}, estimated by a stratified Monte Carlo average over $64$ strata of the diffusion step (one noise draw each). This orders windows by denoising difficulty, and hence by volatility. U-GNN's samples cluster lowest (mean $\approx 0.28$), while the real data lies just to their right (mean $\approx 0.32$). GRW paths form a distinct high-error mode (mean $\approx 0.46$), since memoryless Gaussian increments are hard to denoise for a model that expects temporal and cross-sectional structures. Here, U-GNN nearly coincides with the real data ($W_1{=}0.04$), while GRW sits farther away ($W_1{=}0.14$). Neither score is impartial on its own, but read together, they corroborate that U-GNN better captures the real distribution. 
While U-GNN's score isolates GRW samples as unrealistic, GRW ranks U-GNN above the real data.
That said, the temporal nature of stock data may favor attention-based architectures suited to long-range dependencies. We leave the integration of a graph transformer with our U-GNN backbone to future work.


\subsection{Wireless Resource Allocation}
\label{sec:wireless-experiments}

\begin{table}[!t]
  \centering
  \caption{Power control (WRA) simulation setup.}
  \label{tab:sim-params}
  \begin{tabular}{@{}ll@{}}
    \toprule
    \textbf{Parameter} & \textbf{Value} \\
    \midrule
    \multicolumn{2}{@{}l}{\textit{Network and deployment}}\\
    Tx--rx pairs $N$            & \numNodes \\
    Area side length $R$        & \hspace{-1.5em} $\{\deploymentRultralow, \deploymentRlow, \deploymentRmid, \deploymentRhigh\}$~m \\
    User density $\nu$          & \hspace{-1.5em} $\{
      \pgfmathprintnumber[fixed, precision=1]{\densityultralow},
      \pgfmathprintnumber[fixed, precision=1]{\densitylow},
      \pgfmathprintnumber[fixed, precision=1]{\densitymid},
      \pgfmathprintnumber[fixed, precision=1]{\densityhigh}\}$ pairs/\si{\kilo\meter\squared} \\
    Tx--rx separation          & $\mathrm{Uniform}[\minTxRxSep, \maxTxRxSep]$~m \\
    Min.\ tx--tx spacing        & \minTxTxSep~m \\
    Networks                    & \pgfmathprintnumber[fixed, precision=0]{\numNets} (\numNetsPerDensity~per density) \\
    Train/val/test (per density) & $20/4/8$ ($5{:}1{:}2$) \\
    \midrule
    \multicolumn{2}{@{}l}{\textit{Channel and fading}}\\
    Bandwidth $W$               & \bandwidthInMHz~MHz \\
    Carrier frequency $f_c$     & \carrierFreq~GHz \\
    Noise PSD $N_0$             & \noisePSDIndBmPerHz~dBm/Hz \\
    Max.\ tx power $P_{\max}$    & \PmaxIndBm~dBm (\PmaxInMW~mW) \\
    Shadowing std.\ dev.        & \shadowingStd~dB \\
    \midrule
    \multicolumn{2}{@{}l}{\textit{Allocation timescale}}\\
    Fading step $\delta t$      & \deltaTms~ms \\
    Realizations per network $T$ & \numFadingSteps \\
    Ergodic window $T_0$        & \ergodicWindowSteps~slots ($\slotDurationMs$~ms) \\
    Slots per network $S$       & \numTimeSlots \\
    \bottomrule
  \end{tabular}
\end{table}

We study optimal power control in multi-user interference networks. The setup closely follows our preliminary work on diffusion policies for resource allocation \cite{uslu2025generativediffusionmodelsresource, uslu2026graphsignalwireless} and the related state-augmented formulations of \cite{uslu2025faststateaugmentedlearningwireless}. We summarize it below and collect the simulation parameters in Table~\ref{tab:sim-params}.

\begin{table*}[t]
  \centering
    \caption{\textbf{Ergodic-rate performance on the WRA test split ($\pgfmathprintnumber[fixed, precision=1]{\numTestNets}$ networks, $\pgfmathprintnumber[fixed, precision=1]{\numTestRx}$ tx--rx pairs, $f_{\min}=0.6$\,bits/s/Hz).} \textbf{U-GNN} tracks the expert (PD) policy on the cell-edge percentiles and feasibility, where FP and AP collapse. Per column, the policy performing closest to the reference expert is shown in \textbf{bold} and $\pm$ is one standard deviation across networks.}
  \label{tab:wra-ergodic}
  \setlength{\tabcolsep}{3pt}
  \small
  \begin{tabular*}{\textwidth}{@{\extracolsep{\fill}} l cccc cccc c @{}}
    \toprule
    & \multicolumn{4}{c}{Ergodic rate (bits/s/Hz)\,$\uparrow$}
    & \multicolumn{4}{c}{Gap to expert (\%)\,$\downarrow$}
    & \multirow{2}{*}{Feasible (\%)\,$\uparrow$} \\
    \cmidrule(lr){2-5}\cmidrule(lr){6-9}
    Model
    & p1 & p5 & p10 & mean
    & p1 & p5 & p10 & mean
    & \\
    \midrule
    PD-Expert \emph{(ref.)}
          & $0.66 \pm 0.10$ & $0.82 \pm 0.11$ & $0.93 \pm 0.12$ & $2.84 \pm 0.27$
          & --- & --- & --- & ---
          & $99.1 \pm 1.1$ \\
    \textbf{U-GNN}
          & $\mathbf{0.55} \pm 0.10$ & $\mathbf{0.77} \pm 0.09$ & $\mathbf{0.92} \pm 0.09$ & $2.81 \pm 0.33$
          & \textbf{17.8} & \textbf{5.3} & \textbf{0.9} & 1.1
          & $\mathbf{98.1} \pm 1.1$ \\
    Full power (FP)
          & $0.08 \pm 0.04$ & $0.28 \pm 0.09$ & $0.53 \pm 0.14$ & $3.14 \pm 0.36$
          & 87.9 & 65.4 & 43.0 & $-10.6$
          & $88.0 \pm 2.9$ \\
    Avg.\ power (AP)
          & $0.19 \pm 0.08$ & $0.52 \pm 0.13$ & $0.74 \pm 0.15$ & $\mathbf{2.85} \pm 0.39$
          & 71.8 & 36.8 & 19.7 & $\mathbf{-0.3}$
          & $92.7 \pm 2.7$ \\
    \bottomrule
  \end{tabular*}
  \par\vspace{3pt}
  \begin{minipage}{\textwidth}
    \footnotesize
    \textbf{p$k$}: $k$-th percentile of per-user ergodic rates (worst-served tx--rx pairs). \textbf{Gap}: $100 \times (f^{\text{exp}}-f)/f^{\text{exp}}$ ($>0$: below expert, $<0$: above). \textbf{Feasible}: Percentage of tx--rx pairs meeting $f_{\min}$. FP uses $P_{\max}$ on every link; AP uses each network's mean expert power.
  \end{minipage}
\end{table*}


\smallskip
\noindent \textbf{Setup.} We consider wireless networks of $N = \numNodes$ transmitter--receiver (tx--rx) pairs 
dropped uniformly at random over a square service area. The pairs share a single channel of bandwidth $W$ with noise PSD $N_0$ and per-transmitter power budget $P_{\max}$ (Table~\ref{tab:sim-params}). The channel gain $h_{ij,t} \in \reals_+$ from transmitter $i$ to receiver $j$ at step $t$ combines a static large-scale component with a unit-mean Rayleigh fast-fading component. The large-scale component follows a dual-slope path-loss law with log-normal shadowing and fixes the long-term average gain $h_{ij}$. We collect the long-term and instantaneous gains in $\bbH, \bbH_t \in \reals^{N \times N}_+$, with $[\bbH]_{ij} = h_{ij}$ and $[\bbH_t]_{ij} = h_{ij,t}$.

The policy and the instantaneous gains $\bbH_t$ evolve on separate timescales. The fast-fading process generates $T$ realizations of $\bbH_t$, while the policy emits one allocation per slot and holds it constant over a window of $T_0$ realizations, giving $S = T / T_0$ slots per network (Table~\ref{tab:sim-params}). The network configuration $\bbH$ stays fixed over the operation window and is the only channel state available at allocation time. The policy is conditioned on $\bbH$ via the GSO $\bbS$ and node-state $\bbu$ (defined below), while the instantaneous gains $\bbH_t$ are observed only through the rates they induce.

We model each configuration as a draw $\bbH \sim q_{\bbH}$ from a family of network geometries and large-scale fading. Sweeping the area side length $R$ over four values yields four user densities, each contributing $\numNetsPerDensity$ networks for a dataset of $\pgfmathprintnumber[fixed,precision=0]{\numNets}$ configurations in total, which we split $5{:}1{:}2$ into training, validation, and testing (Table~\ref{tab:sim-params}).

Given an allocation $\bbp$ and a channel realization $\widetilde{\bbH}$, the rate of the receiver $j$ is
\begin{equation} \label{eq:receiver-rate}
  \big[ \widetilde{\bbf}(\bbp, \widetilde{\bbH}) \big]_j
  = \log_2 \! \bigg( 1 + \frac{ [\bbp]_j \, [\widetilde{\bbH}]_{jj} }
    { W N_0 + \sum_{i \neq j} [\bbp]_i \, [\widetilde{\bbH}]_{ij} } \bigg).
\end{equation}
We evaluate a policy by its ergodic rate, estimated by averaging the instantaneous rates over the fading realizations,
\begin{equation} \label{eq:ergodic-receiver-rate}
  \bbf(\bbpi, \bbH)
  = \E_{\bbp, \widetilde{\bbH}}
    \big[\, \widetilde{\bbf}(\bbp, \widetilde{\bbH}) \big]
  \approx \frac{1}{T} \sum_{t=1}^{T}
    \widetilde{\bbf}\big( \bbp_{\lceil t / T_0 \rceil}, \bbH_t \big).
\end{equation}
\noindent In~\eqref{eq:ergodic-receiver-rate}, the expectation is over the policy allocations $\bbp \sim \bbpi(\cdot \cond \bbH)$ and the fast-fading channel $\widetilde{\bbH} \sim q_{\widetilde{\bbH} \vert \bbH}$. The allocation $\bbp_s$ is drawn once per slot $s = \lceil t / T_0 \rceil$ and held fixed across it, while $\bbH_t$ is the channel realized at step $t$.

Given $\bbH$, an optimal policy $\bbpi^\star(\bbH) = \bbpi^\star(\cdot \given \bbH)$ maximizes the ergodic sum-rate subject to a per-user minimum-rate (QoS) requirement $f_{\min} = \fmin$~bits/s/Hz, i.e.,
\begin{subequations} \label{eq:problem}
  \begin{align}
    \bbpi^\star(\bbH) \;\in \;\argmax_{\bbpi} \quad
      & \mathbf{1}_N^\top \, \bbf(\bbpi, \bbH) \label{eq:problem-obj} \\
    \subjectto \quad
      & \bbf(\bbpi, \bbH) \ge f_{\min} \, \mathbf{1}_N. \label{eq:problem-constraint}
  \end{align}
\end{subequations}

The diffusion targets are produced by an \emph{expert primal--dual (PD) algorithm} that solves~\eqref{eq:problem} via Lagrangian dual (sub)gradient ascent. We train a separate expert per density group, parametrized by a shallow GNN, across all networks. We then collect $\numExpertSamples$ allocations per training network from its converged primal iterates, giving $\numTrainSamples$ conditional samples that the diffusion model (U-GNN) learns to reproduce given $(\bbS, \bbu)$. Running the same expert across all splits, including the validation and test networks held out from U-GNN, yields the reference allocations against which we benchmark the U-GNN policy. Since the expert solves each network directly rather than generalizing, it is a near-ideal reference that U-GNN can only approach, so the comparison favors the expert by design. At inference, U-GNN thus samples directly from this \emph{distribution} of near-optimal, feasible allocations rather than re-solving the optimization per network. We refer to \cite{uslu2026graphsignalwireless} for the expert's algorithmic details.

We represent $\bbH$ as a directed, weighted channel graph on $N$ nodes, one per tx--rx pair. A self-loop carries the direct-link gain $h_{ii}$, and an edge carries the interference gain $h_{ij}$, $i \neq j$, which is generally asymmetric, i.e., $h_{ij} \neq h_{ji}$. We sparsify this graph into the GSO $\bbS$ by retaining, for each receiver, its self-loop and the incoming edges from the $m = \topK$ strongest interferers, followed by a log-normalization that compresses the wide dynamic range of the channel gains. We also derive a two-feature node-state vector $\bbu$ that collects each node's direct-link strength and its aggregate incoming interference under full-power transmission. Both $\bbS$ and $\bbu$ are deterministic functions of $\bbH$, so they form the conditioning input $\bbxi = (\bbS, \bbu)$ of Section~\ref{sec:graph-signal-generative-models}, and $q_{\bbH}$ induces its distribution $q_{\bbxi}$. 

The diffusion variable is the rescaled allocation $\bbx_0 = \bbp / P_{\max} - \tfrac{1}{2} \in [-\tfrac{1}{2}, \tfrac{1}{2}]^N$, viewed as a scalar graph signal on $\bbS$. We realize the U-GNN policy, which is trained to imitate the optimal policy $\bbpi^\star$, through the conditional diffusion generator $\pgenstar(\cdot \cond \bbS, \bbu)$ of Section~\ref{sec:graph-signal-generative-models}. That is, given $\bbH$, we sample a batch of $S = \numTraj$ rescaled allocations $\bbx_0 \sim \pgenstar(\cdot \cond \bbS, \bbu)$ and map them to power allocations $\bbp = \mathrm{clip}_{[0, P_{\max}]}\!\big(P_{\max}(\bbx_0 + \tfrac{1}{2})\big)$. We place these allocations in random order as $\{\bbp_s\}_{s=1}^S$ and execute them sequentially, with slot $s$ transmitting $\bbp_s$ over its $T_0$-realization window.

We compare the U-GNN policy against two deterministic baselines that forgo time-sharing. For every slot $s$ and network $\bbH$, full power (FP) sets $\bbp_s = P_{\max}\mathbf{1}_N$, and average power (AP) is the expert's per-network mean allocation, i.e., $\bbp_s(\bbH) \approx \E_{\bbp \sim \bbpi^\star(\cdot \cond \bbH)} \,\big[ \bbp \big]$.

\begin{figure*}[ht!]
  \hspace{-1.75cm}
  \includegraphics[trim=90bp 4.9bp 64bp 76bp, clip, width=1.2\linewidth]{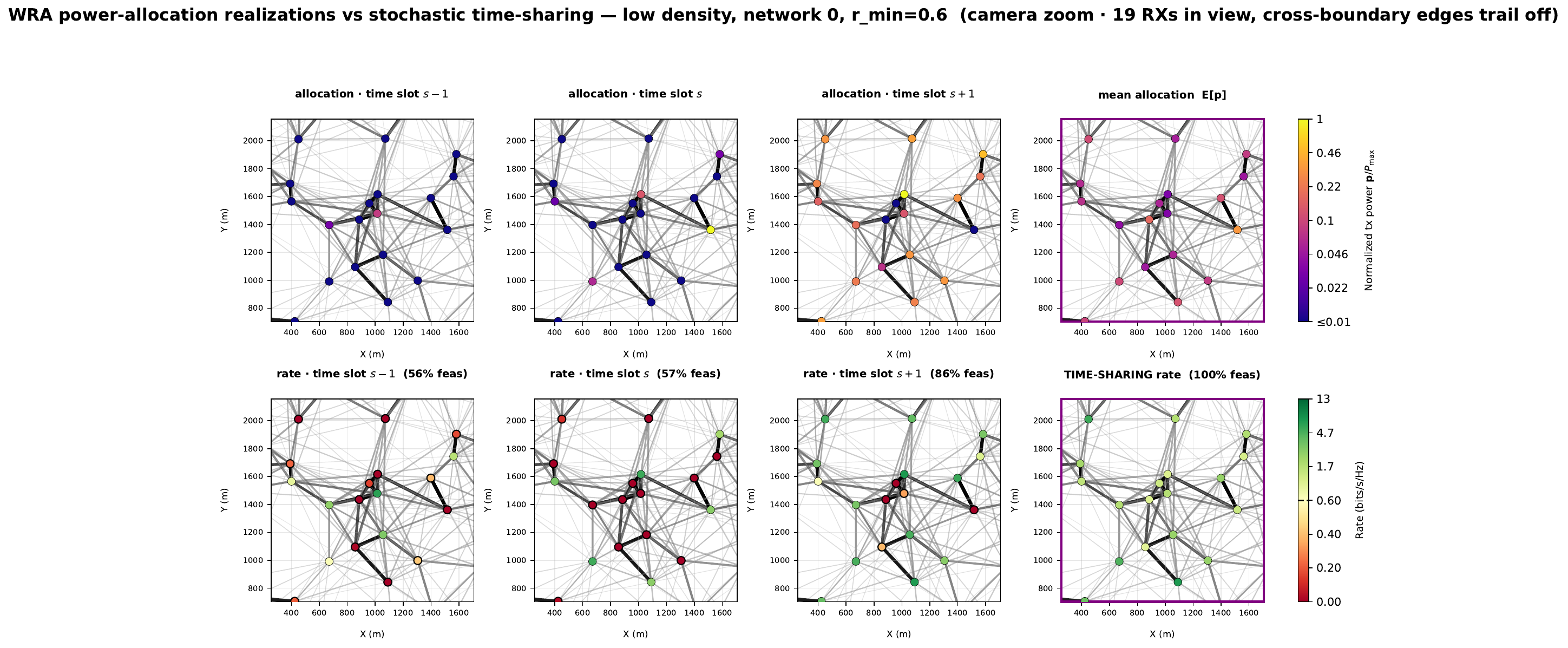}
  \caption{\textbf{U-GNN attains ergodic feasibility through time-sharing.} We zoom in on an example test network, where each node is a tx--rx pair and edges indicate interference strength. The top row shows the per-node normalized transmit power $\bbp/P_{\max}$ (logarithmic color scale) for three consecutive slots of the sequential allocation sampled by the U-GNN policy, alongside the mean allocation $\E[\bbp]$ (rightmost). The bottom row shows the per-receiver rates for each slot (color bar centered at $f_{\min}$) with the per-slot feasibility annotated above, and the time-averaged rate delivered by time-sharing (rightmost). No single slot serves all users ($56$--$86\%$ feasible), yet alternating these allocations across slots mitigates interference and achieves 100\% feasibility for this network. In contrast, the mean allocation rates typically stay infeasible for interference-limited tx--rx pairs.}
\label{fig:wra-time-sharing}

\end{figure*}


\smallskip
\noindent \textbf{Results.} We report in Table~\ref{tab:wra-ergodic} the ergodic-rate performance of the U-GNN policy, expert (PD) policy, and baselines. The U-GNN diffusion policy closely matches the expert across the full distribution. Its 5th- and 10th-percentiles (p$5$, p$10$) rates reach $0.77$ and $0.92$~bits/s/Hz, trailing the expert by only $5.3\%$ and $0.9\%$. The p$5$ rate already exceeds the $f_{\min}=0.6$~bits/s/Hz floor, and only the bottom ${\sim}2\%$ of users fall slightly short. U-GNN's gap to the expert is largest at p$1$, at $17.8\%$ ($0.55$ vs.\ $0.66$~bits/s/Hz), whereas FP and AP baselines trail by $87.9\%$ and $71.8\%$ at this percentile, making U-GNN's degradation minor by comparison. FP attains the highest mean rate at $3.14$~bits/s/Hz while collapsing its p$1$ rate to $0.08$~bits/s/Hz and lowering feasibility to $88.0\%$. U-GNN instead sustains a mean rate of $2.81$~bits/s/Hz, on par with the expert ($2.84$) and AP ($2.85$). AP matches this mean rate but still leaves $7.3\%$ of users below $f_{\min}$, so collapsing the policy to a deterministic mean allocation is not enough.

U-GNN's value rests jointly on its stochastic formulation and generative sampling. The stochastic formulation treats the policy as a distribution over allocations, sustaining feasibility through randomization where the deterministic AP fails. The expert shares this formulation but reaches it through thousands of primal--dual iterations per network, whereas U-GNN bypasses online optimization and attains comparable performance in a single DDIM sampling pass.

\begin{figure}[t!]
    \centering
    \includegraphics[width=\linewidth]{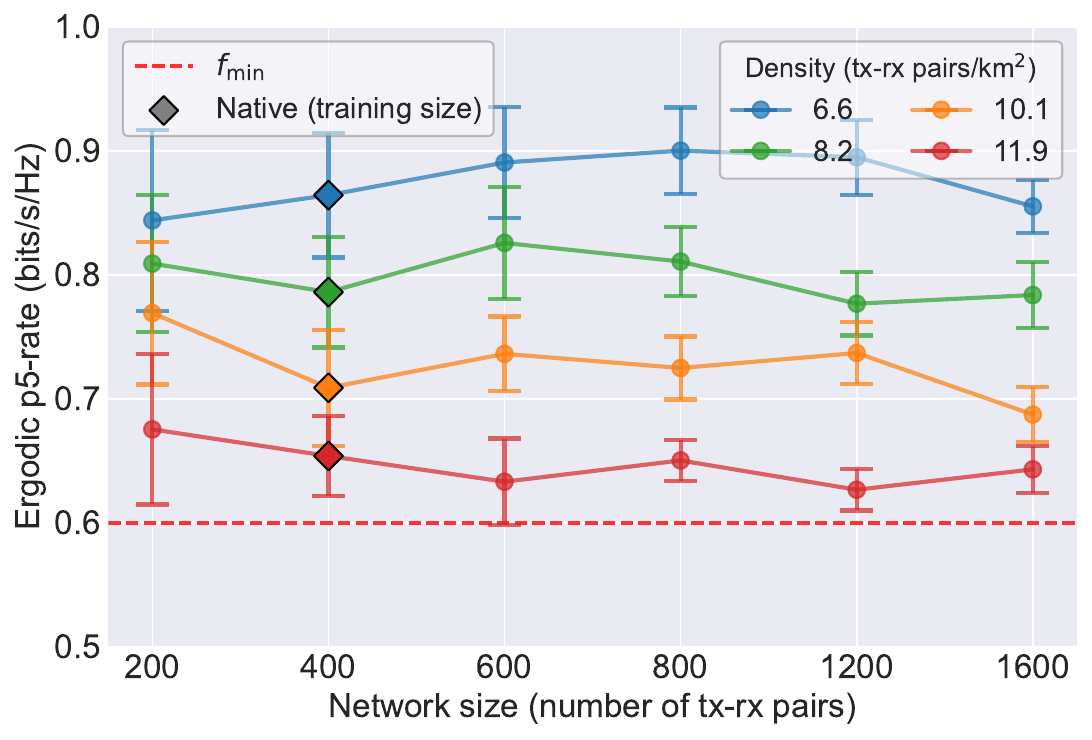}
    \caption{\textbf{Size-transferability of the U-GNN policy.} Ergodic $5$th-percentile (p$5$) rate vs.\ network size for four interference densities (tx--rx pairs/km$^2$) at $f_{\min}=0.6$~bits/s/Hz (red dashed line). Diamonds mark the native training size ($\numRx$ pairs), and all other sizes are evaluated without retraining.
    }
    \label{fig:wra-size-transferability}
\end{figure}

Fig.~\ref{fig:wra-time-sharing} illustrates why the optimal policy must be stochastic and how U-GNN realizes it through \emph{time-sharing}. The optimal (expert) policy is multi-modal, activating a different subset of the mutually interfering tx--rx pairs in each slot. No single allocation serves all users, yet, alternating them across slots renders the time-averaged rate $\bbf(\bbpi^\star, \bbH)$ feasible for every receiver. Crucially, this rate averages the per-slot rates and must not be conflated with the rate of the single mean allocation $\E[\bbp]$, which activates every interferer at once and remains infeasible. This is exactly the failure mode of the AP baseline in Table~\ref{tab:wra-ergodic}.

Fig.~\ref{fig:wra-size-transferability} examines the size-transferability of the U-GNN policy. The model is trained only on networks of the native size
$|\ccalV| = \numRx$ tx--rx pairs (diamonds) with mixed densities. It is then evaluated without retraining on networks ranging from $200$ to $1600$ pairs at similar densities. The ergodic $5$th-percentile rate stays essentially flat and above the $f_{\min}$ floor across this range for every density. 
Denser networks ($11.9$ vs.\ $6.6$ pairs/km$^2$) attain lower tail rates due to stronger
interference, but their ordering and stability are preserved across all sizes. This indicates that the permutation-equivariant GNN backbone
is also scalable. Similar observations hold for the mean and additional tail rates, which are not shown here.

\section{Conclusion \& Future Work}
\label{sec:conclusion}

We proposed a denoising diffusion framework for generating graph signals. We cast this as a denoising process and parametrized it with a U-GNN, a graph-domain adaptation of the U-Net. At its core, a pooling mechanism formalizes down/up-sampling as learned node selection via nested selection matrices and zero-padding. We demonstrated the framework on wireless resource allocation and stock-price forecasting. Future directions include designing graph-aware and latent diffusion processes, and integrating graph transformers into the U-GNN backbone for spatio-temporal tasks.


\bibliographystyle{IEEEbib}
\bibliography{strings,refs}

\clearpage
\newpage
\begin{appendices}

\renewcommand{\thesubsection}{\Alph{section}.\arabic{subsection}}

\setcounter{subsection}{0}

\section{Accelerated DDIM Sampling}
\label{app:accelerated-sampling}
 
The base DDIM sampler \eqref{eq:ddim-sampler} deferred the specification of its noise scale $\sigma_k$. We provide it here, where it also governs the accelerated sampler used in our experiments. A factor $\eta \in [0, 1]$ controls the per-step stochasticity through
\begin{align} \label{eq:ddim-sigma}
    \sigma_k(\eta) = \eta \sqrt{\frac{1 - \alphacumprod_{k-1}}{1 - \alphacumprod_k}}\, \sqrt{1 - \frac{\alphacumprod_k}{\alphacumprod_{k-1}}},
\end{align}
which interpolates between the ancestral DDPM update at $\eta = 1$ and the deterministic DDIM map at $\eta = 0$.
 
Because the DDIM updates in \eqref{eq:ddim-sampler} preserve the marginals of the forward process, we can subsample the denoising steps to accelerate inference without retraining. Concretely, we train $\bbepsilon_{\bbtheta^\star}$ on the full grid of $K$ steps but sample along only an increasing subset $\{\tau_1, \ldots, \tau_{K'}\} \subset \{1, \ldots, K\}$ with $\tau_{K'} = K$ and $K' \ll K$, and we set $\tau_0 \coloneqq 0$. We draw $\bbx_{\tau_{K'}} = \bbx_K \sim \mathcal{N}(\bb0, \bbI)$ and iterate the update \eqref{eq:ddim-sampler} over this sub-grid,
\begin{align} \label{eq:ddim-sampler-accelerated}
    \bbx_{\tau_{k-1}} \!=\! \sqrt{\alphacumprod_{\tau_{k-1}}}\; \widehat{\bbx}_0
    \!+\! \sqrt{1 - \alphacumprod_{\tau_{k-1}} - \sigma_{\tau_k}^2}\; \bbepsilon_{\bbtheta^\star}
    \!+\! \sigma_{\tau_k}\bbw,
\end{align}
for $k = K', \ldots, 1$, and $\bbw \sim \mathcal{N}(\bb0, \bbI)$. We note that both $\widehat{\bbx}_0 = \widehat{\bbx}_0(\bbx_{\tau_k}, \tau_k; \bbxi)$ [cf.~\eqref{eq:noisepred-denoiser-param-equivalence}] and $\bbepsilon_{\bbtheta^\star} = \bbepsilon_{\bbtheta^\star}(\bbx_{\tau_k}, \tau_k; \bbxi)$ are evaluated at the current step $\tau_k$. The noise scale $\sigma_{\tau_k}$ follows from \eqref{eq:ddim-sigma} under the substitution $(k, k{-}1) \mapsto (\tau_k, \tau_{k-1})$. 


%
 
\section{Implementation of Strided Graph Convolutions}
\label{app:strided-impl}
 
A literal evaluation of the strided graph convolution in~\eqref{eq:proposed-graph-convolution-layer} would form the operator $\bbS^\gamma$ and accumulate its powers $(\bbS^\gamma)^k$, $k = 0, \ldots, K$. For $\gamma > 1$, this densifies $\bbS$ and forfeits its sparsity, much like the reduced GSOs $\bbS^{(k)}$ of~\eqref{eq:reduced-GSOs} do. We instead exploit $(\bbS^\gamma)^k = \bbS^{\gamma k}$ and evaluate the layer through $\gamma K$ successive sparse products with the unit shift $\bbS$, tapping the signal every $\gamma$ shifts. 

Starting from the lifted signal $\widetilde{\bbZ}_{0} = \bbD^\top \bbZ_{\ell-1}$, we form $\widetilde{\bbZ}_{j} = \bbS\, \widetilde{\bbZ}_{j-1}$ for $j = 1, \ldots, \gamma K$ and accumulate the tapped terms $\widetilde{\bbZ}_{\gamma k}\, \bbTheta_{\ell,k}$, $k = 0, \ldots, K$. We then apply the pointwise nonlinearity $\sigma_{\ell}$ and the pooling $\bbD$. The untapped shifts, $j \notin \{0, \gamma, \ldots, \gamma K\}$, only advance the walk to the next tapped hop. Algorithm~\ref{alg:strided-conv} summarizes the procedure.
 
\begin{algorithm}[t]
\caption{Strided graph convolutions via iterated unit shifts}
\label{alg:strided-conv}
\begin{algorithmic}[1]
\Require Input signal $\bbZ_{\ell-1} \in \reals^{N' \times F_{\ell-1}}$; shift $\bbS$; selection $\bbD$; taps $\{\bbTheta_{\ell,k}\}_{k=0}^{K}$; stride $\gamma$; pointwise nonlinearity $\sigma_{\ell}$
\Ensure Output signal $\bbZ_{\ell} \in \reals^{N' \times F_{\ell}}$
    \State $\widetilde{\bbZ}_{0} \gets \bbD^\top \bbZ_{\ell-1}$ \Comment{lift to the full vertex set}
    \State $\bbA \gets \widetilde{\bbZ}_{0}\, \bbTheta_{\ell,0}$ \Comment{tap $k = 0$}
    \For{$j = 1, \ldots, \gamma K$}
        \State $\widetilde{\bbZ}_{j} \gets \bbS\, \widetilde{\bbZ}_{j-1}$ \Comment{one sparse shift; a single $N \times F$ buffer suffices}
        \If{$\gamma \mid j$}
            \State $\bbA \gets \bbA + \widetilde{\bbZ}_{j}\, \bbTheta_{\ell,\, j/\gamma}$ \Comment{tap every $\gamma$ shifts}
        \EndIf
    \EndFor
    \State $\bbZ_{\ell} \gets \bbD\, \sigma_{\ell}(\bbA)$ \Comment{nonlinearity, then reduce to $\Omega$}
    \State \Return $\bbZ_{\ell}$
\end{algorithmic}
\end{algorithm}
 
Notably, we never materialize the reduced GSOs. The shift $\bbS$ is a sparse weighted aggregation, while the unpooling $\bbD^\top$ and pooling $\bbD$ are index scatter and gather operations. Only the running buffer $\widetilde{\bbZ}_{j}$ and one accumulator are kept, each of size $N \times F$, so each layer runs in roughly $\mathcal{O}(\gamma K\, |\ccalE|\, F)$ aggregation time and $\mathcal{O}(N F)$ memory, where $|\ccalE|$ is the number of edges of $\ccalG$. Forming the dense $\bbS^\gamma$ explicitly would instead cost $\mathcal{O}(N^2)$ memory.
The filter keeps $K+1$ taps for any $\gamma$, so the stride adds shifts but no parameters.
 
The stride enters only as a multiplier on the shift count and is tied to the pooling ratio. When the GNN module sits in a depth-$b$ block of the U-GNN, it instantiates~\eqref{eq:proposed-graph-convolution-layer} with $\bbD = \bbD_b$, and all of its layers share the stride $\gamma_b = \min \{ \lfloor \sqrt{\bar\rho_b}\, \rfloor,\, \gamma_{\max} \}$, where $\bar\rho_b = \prod_{i=1}^{b} \rho_i = N / N_b$ is the cumulative down-sampling factor [cf.~\eqref{eq:sampling-matrix}]. This is the rule $\gamma = \lfloor \sqrt{\rho}\, \rfloor$ of Remark~\ref{remark:stride} applied at depth $b$ with $\rho = \bar\rho_b$ and capped at $\gamma_{\max}$, so the total reach $\gamma_b K$ does not span the whole graph and we avoid oversmoothing.

\section{Straight-Through Estimation and Stochastic Top-K Selection}
\label{sec:ste}
To train the score head $\bbPsi_b$ by backpropagation, we apply a
straight-through estimator (STE)~\cite{bengio2013estimating} to the binary selection mask. That is, the forward pass
selects a hard mask, while gradients reach $\bbv_b$ through a differentiable sigmoid surrogate. Moreover, we employ a Gumbel-perturbed Top-K rule to encourage exploration during training.

\smallskip
\noindent \textbf{Stochastic selection.} Let $\bbh_b \in \{0,1\}^{N_b}$ be the Top-K
indicator, with $[\bbh_b]_n = \mathbbm{1}\{n \in \mathcal{S}_b\}$. At inference,
$\mathcal{S}_b$ is the deterministic Top-K of~\eqref{eq:topk}. During training, we draw $\mathcal{S}_b$ stochastically to encourage exploration of the discrete selection. Given i.i.d.\ standard Gumbel variates
$[\bbg_b]_n = -\log(-\log u_n)$, $u_n \sim \mathrm{Uniform}(0,1)$, and an
exploration scale $\varepsilon \in \reals_{+}$, we perturb the scores before
ranking,
\begin{align}
    \mathcal{S}_b = \mathrm{TopK}(\bbv_b + \varepsilon\,\bbg_b,\, N_{b+1}).
    \label{eq:gumbel-topk}
\end{align}
Note that Top-K is invariant to positive scaling, i.e.,
$\mathrm{TopK}(\bbv_b + \varepsilon\bbg_b,\, N_{b+1})
= \mathrm{TopK}(\bbv_b/\varepsilon + \bbg_b,\, N_{b+1})$. By the Gumbel-Top-K
trick~\cite{kool2019stochastic}, \eqref{eq:gumbel-topk} therefore draws $\mathcal{S}_b$ without replacement with probabilities proportional to $e^{[\bbv_b]_n/\varepsilon}$. The scale $\varepsilon$ acts as a sampling temperature, and as
$\varepsilon \to 0$, the sample concentrates on the deterministic Top-K, which we recover at inference when the perturbation vanishes at $\varepsilon = 0$.

\smallskip
\noindent \textbf{Straight-through gradient.} For the backward pass, we form a surrogate from
the \emph{unperturbed} scores. Given a temperature $\tau \in \reals_{+}$ and the
logistic sigmoid $\varsigma(\cdot)$, we set
$\widetilde{\bbh}_b = \varsigma(\bbv_b/\tau) \in (0,1)^{N_b}$. The training mask is
\begin{align}
    \bbm_b = \bbh_b + \widetilde{\bbh}_b - \mathrm{sg}(\widetilde{\bbh}_b),
    \label{eq:ste}
\end{align}
where $\mathrm{sg}(\cdot)$ is the stop-gradient operator. The soft terms cancel in
value, thus $\bbm_b = \bbh_b$ in the forward pass. In the backward pass,
$\mathrm{sg}(\tilde{\bbh}_b)$ is constant, so the gradient reaches $\bbv_b$ only
through $\tilde{\bbh}_b$. We note that because the surrogate uses the unperturbed scores, the Gumbel noise affects only which nodes enter $\mathcal{S}_b$, not the surrogate gradient at any node.

\smallskip
\noindent \textbf{Gated down-sampling.} We insert this mask into the encoder down-sampling.
During training, the depth-$(b{+}1)$ input
$\bbV_{b+1} = \bbC_{b+1}\,\sigenc_b$ of~\eqref{eq:block-input-fusion} becomes
\begin{align}
    \bbV_{b+1} = \bbC_{b+1}\big(\bbm_b \odot \sigenc_b\big),
    \qquad b = 1, \ldots, B-1,
    \label{eq:downsampled-signal}
\end{align}
where $\odot$ scales each row of $\sigenc_b$ by the corresponding entry of
$\bbm_b$. Since $\bbm_b = \bbh_b$ in the forward pass, the selected rows pass
through unchanged, and $\bbC_{b+1}$ discards the rest. At inference, the STE branch
is dropped ($\bbm_b = \bbh_b$), and~\eqref{eq:downsampled-signal} reverts to the
plain down-sampling via $\bbC_{b+1}\,\sigenc_b$.

\smallskip 
\noindent \textbf{Schedules.} We anneal both $\tau$ and $\varepsilon$ linearly over training. Let
$t$ and $T$ denote the current and total number of training epochs, with $T = \maxEpochs$ as reported in Table~\ref{tab:impl}. The schedule uses a warm-up of $T_{\mathrm{w}}$ epochs followed by
an anneal of $T_{\mathrm{a}}$ epochs,
\begin{equation}
    \begin{aligned}
        \tau(t) &= \tau_0 + r(t)\,(\tau_{\min} - \tau_0), \\
        \varepsilon(t) &= \varepsilon_0 + r(t)\,(\varepsilon_{\min} - \varepsilon_0),
    \end{aligned}
    \label{eq:sel-schedule}
\end{equation}
\noindent where $r(t) = \min \{ \max \{(t - T_{\mathrm{w}})/T_{\mathrm{a}}, \, 0 \}, \, 1 \}$. We set $T_{\mathrm{w}} = 0.02\,T$, $T_{\mathrm{a}} = 0.73\,T$,
$(\tau_0, \tau_{\min}) = (1, 0.5)$, and
$(\varepsilon_0, \varepsilon_{\min}) = (1, 0)$. This way, both $\varepsilon$ and $\tau$ reach their floors at epoch
$T_{\mathrm{w}} + T_{\mathrm{a}} = 0.75\,T$, after which selection is (approximately) deterministic, and the surrogate sharpness is fixed. These values worked well in our experiments, but we found the results to be robust, as any choice that explores broadly early and becomes deterministic well before training ends gave comparable performance.

We remark that the STE in~\eqref{eq:ste} supplies a surrogate gradient for the binary mask, and not a differentiable relaxation of the hard Top-K rule. In
the forward pass, $\bbC_{b+1}$ removes the unselected rows, so each training iteration passes diffusion-loss gradients only to the score entries of the nodes
in $\mathcal{S}_b$. The head nonetheless learns a global ranking for two reasons. First, while $\varepsilon > 0$, the stochastic selection varies $\mathcal{S}_b$
across iterations, so the gradient reaches a broader set of nodes over training. Second, the head shares its parameters across all candidate nodes, so each update
reshapes the scoring function at every node in subsequent passes.

\begin{figure*}[ht!]
  \centering
  \includegraphics[trim=6bp 159bp 6bp 4bp, clip, width=\textwidth]{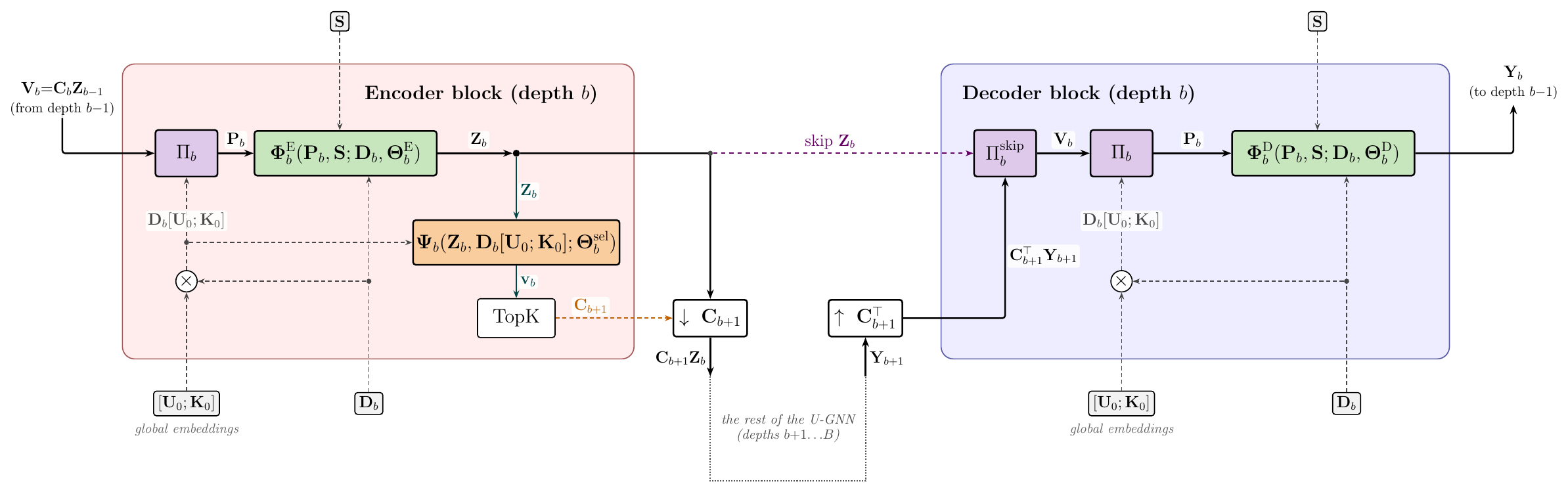}
  \caption{\textbf{Block interface of a U-GNN encoder--decoder pair.} 
  \emph{Encoder (left):} The block input $\bbV_b{=}\bbC_b\,\sigenc_{b-1}$ passes through
  the fusion layer $\projlayer_b$ and the GNN module $\bbPhi^{\mathrm E}_b$ to produce
  $\sigenc_b$. This feature continues along the skip and also feeds the learned selector
  $\bbPsi_b$. A Top-K readout of the selector scores then yields the selections
  $\bbC_{b+1}$.
  \emph{Junction (centre):} The down-sampled $\bbC_{b+1}\sigenc_b$ descends into the
  elided deeper levels $b{+}1,\dots,B$. The returning $\sigdec_{b+1}$ is up-sampled by
  $\bbC_{b+1}^{\top}$.
  \emph{Decoder (right):} The skip-projection $\projlayerskip_b$ combines the up-sampled signal with the encoder skip $\sigenc_b$. The result passes through $\projlayer_b$ and the GNN module $\bbPhi^{\mathrm D}_b$, producing $\sigdec_b$ for depth $b{-}1$.
  In every block, the global embeddings $[\bbU_0;\bbK_0]$, which are restricted to the depth-$b$
  active nodes as $\bbD_b[\bbU_0;\bbK_0]$, enter $\projlayer_b$ and $\bbPsi_b$. The composite $\bbD_b$ also parametrizes the GNN modules $\bbPhi^{\mathrm E}_b$ and
  $\bbPhi^{\mathrm D}_b$, which define graph convolutions over the GSO $\bbS$.}
  \label{fig:ugnn-block}
\end{figure*}

\section{Classical Sum/Max-Pooling over Graph Neighborhoods}
\label{app:sec:pooling}
 
CNNs with pooling couple neighborhood aggregation and spatial down-sampling into a single operation that summarizes each pixel window, e.g., by its average or maximum, at a reduced resolution. This coupling exploits the intrinsic geometry of the regular grid. For a general graph, the GSO $\bbS$ supplies the analogous structural information, which we use to define pooling neighborhoods.
 
Our design decouples these two operations. The strided graph convolution in~\eqref{eq:proposed-graph-convolution-layer} already returns a nonlinear summary of each node's neighborhood. In reduced indices, this is the $K$-hop reach of $\{\bbS^{(k)}\}_{k=0}^{K}$, and in the lifted domain, it is the $\gamma K$-hop reach of $\bbS$ after $\bbD_b^\top$ lifts the signal and before $\bbD_b$ reduces it back [cf.~\eqref{eq:reduced-GSOs}]. The learned selectors $\bbC_b$ in turn perform the node-space down-sampling [cf.~\eqref{eq:topk}]. An explicit sum- or max-pooling over graph neighborhoods is therefore optional rather than essential. We describe it next for completeness.
 
\smallskip
\noindent\textbf{Pooling over the full graph.}
For a given $\bbS$ and pooling range $\kappa \in \mathbb{Z}^+$, we define the binary \emph{reachability matrix}
\begin{align} \label{eq:reachability-matrix}
    \big[ \bbR(\bbS) \big]_{n,m}
    \coloneqq
    \mathbbm{1} \Big\{ \big[ \bbS^k \big]_{n,m} \neq 0 \ \text{for some}\ 0 \leq k \leq \kappa \Big\}.
\end{align}
Note that each node reaches itself. From $\bbR = \bbR(\bbS)$ we read off, for each node $n \in \ccalV$, the index set
\begin{align} \label{eq:reach-index-set}
    \bbn(n) \coloneqq \left\{ m \in \ccalV \colon \big[ \bbR \big]_{n,m} = 1 \right\}
\end{align}
of nodes aggregated at $n$. Let $\varphi \colon \reals^{\cdot \times F} \to \reals^{F}$ be a pooling map that is invariant to permutations of its rows (the pooled nodes) and acts per feature, e.g., $[\varphi_{\mathrm{sum}}(\bbX)]_{f} = \sum_{i} [\bbX]_{i,f}$ for sum-pooling and $[\varphi_{\max}(\bbX)]_{f} = \max_{i} [\bbX]_{i,f}$ for max-pooling. In analogy with the graph-filter notation $\bbH(\bbS)\bbX$, we write $\bbR(\bbS)\bbX$ for the pooled signal,
\begin{align} \label{eq:pool-operator}
    \big[ \bbR(\bbS) \, \bbX \big]_n
    \;=\; \varphi \big( \big[ \bbX \big]_{\bbn(n)} \big), \quad n \in \ccalV.
\end{align}
Note that for sum-pooling, this operator form coincides with the matrix product $\bbR(\bbS)\bbX$.

\smallskip
\noindent\textbf{Pooling at reduced resolutions.}
Pooling at the resolutions of the U-GNN follows the reduced-GSO construction. We define the node-reduced counterpart of $\bbR$,
\begin{align} \label{eq:reduced-reachability-matrix}
    \bbR_b(\bbS) = \bbD_b \, \bbR(\bbS) \, \bbD_b^\top \in \{0,1\}^{N_b \times N_b},
\end{align}
which intersects the $\kappa_b$-reach of $\bbS$ with the active set $\Omega_b$. Like the reduced GSOs in~\eqref{eq:reduced-GSOs}, $\bbR_b$ is indexed by the local labels $[N_b]$. Also, the range $\kappa_b$ is left implicit at depth $b$, just as the stride $\gamma_b$ is in~\eqref{eq:proposed-graph-convolution-layer}. We define the set
\begin{align} \label{eq:reduced-reach-index-set}
    \bbn_b(n) \coloneqq \left\{ m \in [N_b] \colon \big[ \bbR_b \big]_{n,m} = 1 \right\},
\end{align}
which collects the active nodes within $\kappa_b$ hops of $n$ on $\bbS$, including $n$ itself. We then pool a down-sampled signal $\bbZ \in \reals^{N_b \times F}$ as
\begin{align} \label{eq:reduced-pool-operator}
    \big[ \bbR_b(\bbS) \, \bbZ \big]_n
    = \varphi \big( \big[ \bbZ \big]_{\bbn_b(n)} \big), \quad n \in [N_b].
\end{align}
\noindent In~\eqref{eq:reduced-pool-operator}, $\bbR_b(\bbS)\bbZ$ admits a lift--pool--reduce form mirroring that of the strided convolutions. One lifts $\bbZ$ to the full vertex set, pools over the $\kappa_b$-hop neighborhoods of $\bbS$, and reduces back to $\Omega_b$ through $\bbD_b$. The lift assigns each inactive node the identity element of $\varphi$, so that it leaves the pooled output unchanged. This element is $0$ for sum-pooling, for which the plain $\bbD_b^\top$ suffices, and $-\infty$ (a large negative constant in practice) for max-pooling, supplied by a modified lift $\widetilde{\bbD}_b^\top$.

 
\smallskip
\noindent\textbf{Optional placement in the U-GNN.}
At each encoder depth $b = 1, \ldots, B-1$, the encoder output $\sigenc_b$ is first pooled as $\bbR_b \sigenc_b$ and then down-sampled by $\bbC_{b+1} \bbR_b \sigenc_b$, which feeds the block at depth $b+1$ [cf.~\eqref{eq:encoding-path}]. The node selectors use the pooled features as well, so we replace $\sigenc_b$ with $\bbR_b \sigenc_b$ in~\eqref{eq:node-selector-block}. Both stages fold into the definition of the encoding GNN module. Note that the bottleneck at depth $B$ performs no node-space down-sampling or pooling.

The case for explicit pooling is weak in our design since the graph convolutional layers in~\eqref{eq:proposed-graph-convolution-layer} and selector heads in~\eqref{eq:node-selector-block} already provide nonlinear neighborhood aggregation and learned down-sampling, respectively. In our experiments, max-pooling added computational overhead without a noticeable gain. Hence, we adopt~\eqref{eq:proposed-graph-convolution-layer} as the default GNN operation.

\begin{remark} \label{remark:max-compose}
Max-pooling admits a compositional shortcut. By associativity and the telescoping of iterated neighborhoods, the $\kappa$-reach maximum equals $\kappa$ applications of the unit-reach maximum. This avoids forming $\bbR(\bbS)$ explicitly and reduces the cost to $\kappa$ sparse \texttt{scatter\_max} calls. The same property holds in the reduced domain.
\end{remark}
 
\begin{remark}
For a regular graph with uniform down-sampling, a natural CNN analog sets the pooling range $\kappa_b = \kappa \lceil N / N_b \rceil$, proportional to the effective down-sampling at depth $b$, so that pooling regions cover comparable numbers of active nodes across resolutions. For an arbitrary graph, the tension that governs the stride $\gamma$ reappears [cf.~Remark~\ref{remark:stride}]. Namely, a too small $\kappa_b$ collapses $\bbn_b(n)$ to $\{n\}$ and reduces pooling to the identity, whereas a too large $\kappa_b$ pools broadly over $\Omega_b$ and may oversmooth the representation.
\end{remark}

\section{U-GNN Denoiser and Diffusion Configuration}
\label{app:impl}

This appendix expands Table~\ref{tab:impl}. In particular, Appendix~\ref{app:impl-shared} details the denoiser, diffusion process, and training shared across the two tasks, and Appendix~\ref{app:impl-apps} specifies the signal and conditioning interface particular to each.

\subsection{Shared Denoiser, Diffusion, and Training}
\label{app:impl-shared}

\smallskip
\noindent\textbf{Backbone.}
The denoiser $\bbepsilon_{\bbtheta}$ is a U-GNN spanning $B{=}4$ node resolutions. Each of its $\numPoolLevels$ down-sampling levels reduces the active node set by a factor
$\poolFactor$, for an $\poolFactor^{\numPoolLevels}{=}8\times$ reduction overall and four resolutions of $N$, $N/2$, $N/4$, and $N/8$ nodes. Each reduction is realized by a learned node-selection pooling operator, made differentiable with a straight-through estimator (Appendix~\ref{sec:ste}). To counteract the increasing sparsity of the coarsened graphs, the graph convolutions are strided, capped at $\gamma_{\max}{=}\maxStride$ (Appendix~\ref{app:strided-impl}). The feature (channel) width is uniform at $\baseCh$ channels across all levels. 

Each encoding and decoding level applies $\gnnLayers$ GNN layers of order $K{=}\gnnHops$ over powers of the augmented adjacency
$\bbI+\hat{\bbA}$, with channel-wise layer normalization (LayerNorm), a rectified linear unit (ReLU) nonlinearity, and dropout $\dropoutRate$; these layers are strided variants of the topology-adaptive graph convolution (TAGConv), following its
\href{https://pytorch-geometric.readthedocs.io/en/latest/generated/torch_geometric.nn.conv.TAGConv.html}{\texttt{PyG} implementation}.
A $\numBottleneck$-layer bottleneck processes the coarsest summary, and the decoder path
restores full resolution while concatenating the encoder skip features at each level.
The output head applies LayerNorm, a sigmoid linear unit (SiLU) nonlinearity, and a linear map. Fig.~\ref{fig:ugnn-block} complements Fig.~\ref{fig:ugnn} and shows a detailed schematic of one encoder--decoder block pair at depth $b$.

\smallskip
\noindent\textbf{Conditioning and fusion.}
The diffusion step $k$ is mapped to a $\timeEmbDim$-dimensional sinusoidal embedding
followed by a multilayer perceptron (MLP), and the node conditioning to a
$\condEmbDim$-dimensional representation. Both are projected to the level width
$\baseCh$ at every level, where the signal path and the projected step embedding are
concatenated and linearly mapped back to $\baseCh$ channels ($128{\to}64$). The
conditioning is then merged in a task-specific way: by concatenation into the same
projection for WRA, so that it maps $192$ channels down to $64$ over the signal, step, and
conditioning; and by two-head cross-attention for S\&P~500---queries from the signal
path, keys and values from the conditioning context---which leaves the projection from 
$128$ channels down to $64$.

\smallskip
\noindent\textbf{Diffusion.}
The forward process applies a linear $\beta$ schedule from $\betaStart$ to $\betaEnd$ over $\numDiffSteps$ steps. At inference, we sample using $5\times$-accelerated denoising diffusion implicit models (DDIM) with $\numSampleSteps$ steps and stochasticity parameter $\eta{=}\ddimEta$
(Appendix~\ref{app:accelerated-sampling}). We draw $\numTraj$ samples per conditioning input to form the predictive ensemble.

\smallskip
\noindent\textbf{Optimization.}
We train for $\maxEpochs$ epochs with AdamW (learning rate $\learnRate$,
$(\beta_1,\beta_2){=}\adamBetas$, weight decay $\weightDecay$), gradient-norm clipping
at $\gradClip$, and automatic mixed precision (AMP). The learning rate follows a per-step cosine schedule with a linear warm-up over the first $\warmupFrac$ of training, then a
cosine decay to a floor of $\minLrFrac$ of the peak over $\decayFrac$ of training, after which it is held constant. The node-selector temperature and exploration noise are
annealed over roughly the first $75\%$ of training, moving the selector from soft selection toward nearly hard selection (Appendix~\ref{sec:ste}). We retain the best checkpoint under a task-specific composite validation criterion. Both models were trained on a single NVIDIA RTX~3090. The S\&P~500 model completed training in ${\approx}58.5$~GPU hours, and the WRA model did so in ${\approx}33$~GPU hours.

\subsection{Application-Specific Knobs}
\label{app:impl-apps}

\smallskip
\noindent\textbf{Stock-price forecasting (S\&P~500).}
The diffused signal is the future log-return trajectory $\bbx_0\in\reals^{N\times\Tp}$
over the $N{=}\numStocks$ stocks. The pooling cascade coarsens the node set as
$\numStocks{\to}234{\to}117{\to}58$ across the four levels. Because returns are
unbounded, the predicted clean signal is left unclipped. 

The conditioning comprises
$\numFeat$ market features over the past $\Th$-day window. A per-node temporal encoder
summarizes this window and projects the $\Th$ history steps onto the $\Tp$ horizon
steps, forming the per-level context that the cross-attention fusion above consumes.
This encoder interleaves dilated 1-D convolutions with two-head temporal
self-attention. A lightweight in-block temporal mixer further couples the $\Tp$ horizon
steps. The target channel is additionally normalized by reversible instance
normalization (RevIN)~\cite{kim2022reversible}, using statistics from the conditioning window. RevIN uses a
blend weight of $\revinBlend$ and a scale correction of $\revinAlpha$. 

The model has
$\numParamsStock\mathrm{M}$ trainable parameters and is trained on batches of $\batchStock$ sliding windows. We retain the checkpoint that minimizes a composite of the validation noise loss, its train--validation gap, and
continuous ranked probability score (CRPS). This checkpoint is reached near epoch $4500$.

\smallskip
\noindent\textbf{Wireless resource allocation (WRA).}
The diffused signal is the centered allocation
$\bbx_0=\bbp/P_{\max}-\tfrac{1}{2}\in[-\tfrac{1}{2},\tfrac{1}{2}]^{N}$ over the
$N{=}\numNodes$ transmitter--receiver (tx--rx) pairs. The pooling cascade coarsens the node set
as $\numNodes{\to}200{\to}100{\to}50$. The powers are recovered as $\bbp=\mathrm{clip}_{[0,P_{\max}]}\!\big(P_{\max}(\bbx_0+\tfrac{1}{2})\big)$. In contrast to the stock model, the
allocation is a single snapshot with no time axis, so $T{=}1$. Thus, the temporal encoder and
in-block mixer are therefore disabled, and RevIN is not used. 

The $\condChannelsWra$
per-node conditioning features are concatenated into the level fusion described above.
These features are the direct-link gain and the aggregate cross-link interference. Because the support is bounded, the predicted clean signal is clipped to the valid power range at each reverse step. 

Target allocations are supplied by a primal--dual expert. The
expert is a shallow GNN with $\pdLayers$ layers, hidden width $\pdHidden$, and
$K{=}\pdHops$ hops, trained per density group with dual-ascent step $\pdDualStep$. As
supervision for the U-GNN, we collect $\numExpertSamples$ allocations per training
network from its converged primal iterates. This yields $\numTrainSamples$ samples in
total. Running the same expert on the held-out networks yields the reference
allocations used for evaluation. Theoretical and algorithmic details appear
in~\cite{uslu2026graphsignalwireless}. 

The model has $\numParamsWra\mathrm{M}$
trainable parameters and is trained on batches of $\batchWra$ network graph and resource allocation signal pairs. We retain the checkpoint that minimizes a composite of the mean rate-feasibility gap and the $1\%$ and $5\%$ rate gaps, averaged over the four density
groups. 
This checkpoint is reached at epoch $1600$.




\end{appendices}

\end{document}